\documentclass{article}

\usepackage[preprint]{neurips_2026}

\usepackage[utf8]{inputenc} 
\usepackage[T1]{fontenc}    
\usepackage{hyperref}       
\usepackage{url}            
\usepackage{booktabs}       
\usepackage{amsfonts}       
\usepackage{nicefrac}       
\usepackage{microtype}      
\usepackage{xcolor}         

\usepackage{cuted}
\usepackage{graphicx}
\usepackage{booktabs} 
\usepackage{amssymb}  
\usepackage{makecell} 
\usepackage{amsmath}
\usepackage{subfigure}
\usepackage[most]{tcolorbox}
\usepackage[normalem]{ulem}
\usepackage{algorithm}
\usepackage{algpseudocode}

\usepackage{multirow}
\newcommand{\cmark}{\checkmark}
\newcommand{\xmark}{$\times$}
\usepackage{tabularx}
\usepackage{wrapfig}

\title{DailyReport: An Open-ended Benchmark for Evaluating Search Agents on Daily Search Tasks}

\author{
    \textbf{Jingxuan Han$^{1,*,\ddagger}$, Wei Liu$^{2,*}$, Mingyang Zhu$^{2,*}$, Youpeng Wang$^{1,\ddagger}$, Ziwen Wang$^{2}$,} \\
    \textbf{Lin Qiu$^{2,\dagger}$, Xuezhi Cao$^{2}$, Xunliang Cai$^{2}$, Zheren Fu$^{1}$, Licheng Zhang$^{1}$, Zhendong Mao$^{1,\S}$} \\
    $^1$University of Science and Technology of China \\
    $^2$Meituan \\
    \texttt{\{hjx999222, wyp220517\}@mail.ustc.edu.cn} \\
    \texttt{\{liuwei304, zhumingyang09\}@meituan.com} \\
    $^*$Equal contribution. $^\dagger$Project leader. $^\S$Corresponding author. \\ $^\ddagger$Work was done during their internship.
}

\begin{document}

\maketitle

\begin{abstract}
Search Agents (SAs) typically leverage large language models (LLMs) to support complex information-seeking tasks by autonomously exploring web sources and synthesizing information into comprehensive responses. For SAs evaluation, prior benchmarks mainly focus on specialized tasks that are unlikely to arise in real-world user scenarios. Moreover, their reliance on coarse task-level rubrics often limits evaluation interpretability. To bridge this gap, we introduce \textbf{DailyReport}, an open-ended benchmark to evaluate SA capabilities on daily search tasks. It contains 150 open-ended tasks with 3,546 associated rubrics, capturing widely discussed and timely information demands of real-world users. Each task is decomposed into subtasks and evaluated with cascade rubrics across disentangled dimensions. Through cascade performance attribution and user-centric aggregation, we derive highly interpretable scores for each dimension, along with a user preference score. Our results on 17 agentic systems show that current systems still fall short of users' expectations. To facilitate future research, our dataset and code are made publicly available at \url{https://github.com/AGI-Eval-Official/DailyReport}.
\end{abstract}

\section{Introduction}

With the rise of open-domain web agents, information seeking is moving from traditional keyword retrieval to agentic research. Search Agents (SAs) have therefore emerged to address users' information needs through extensive web exploration and long-horizon reasoning \cite{huang2025deep}. These agents can explore hundreds of web sources and synthesize heterogeneous information into comprehensive responses \cite{wang2025liveresearchbench}. As these agentic systems become increasingly capable, it is essential to evaluate their ability to conduct large-scale information gathering and reasoning.

Recently, several benchmarks have been introduced to evaluate the SAs \cite{fan2025understanding}.
For task construction, most works \cite{wei2025browsecomp, du2025deepresearch,abaskohi2025drbench,sharma2025researchrubrics} rely on domain experts to construct specialized research tasks. These tasks mainly assess agents on overprocessed or professional questions within specific fields, which are unlikely to arise in real-world scenarios. 
Moreover, their static design fails to capture evolving real-world information needs and raises concerns about potential data contamination.
For evaluation, existing studies \cite{xu2025researcherbench, li2026deepresearch} generally define task-level rubrics over coarse-grained dimensions and aggregate their scores linearly. This often undermines evaluation interpretability and fails to quantify performance from the user perspective.

\begin{figure*}[t] 
\centering 
\includegraphics[width=0.97\textwidth]{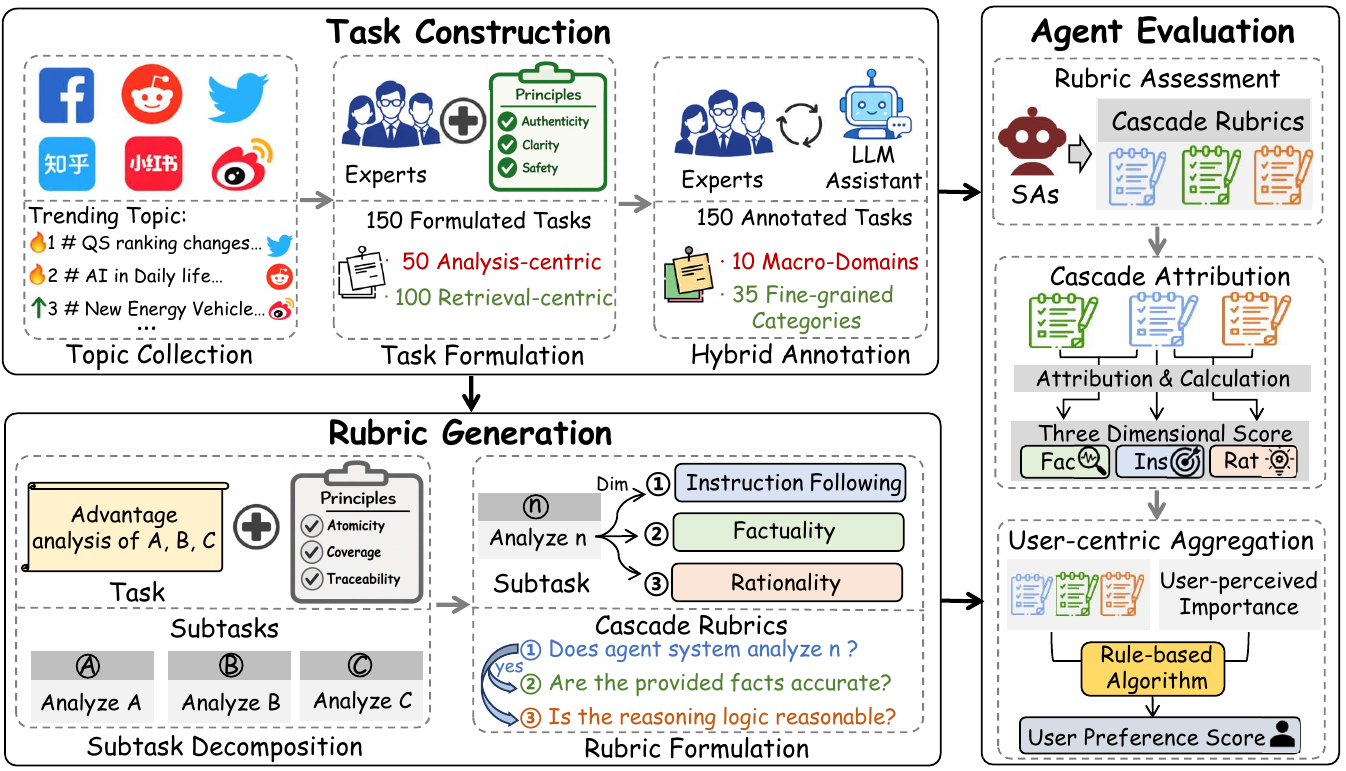} 
\caption{DailyReport structure. We construct daily search tasks and cascade rubrics for evaluating search agents.}
\label{Fig.intro} 
\end{figure*}

In this work, we propose \textbf{DailyReport}, an open-ended benchmark to evaluate SAs on \textbf{daily search tasks}. Unlike previous benchmarks centered on specialized domain problems, DailyReport primarily evaluates whether agents can reliably satisfy everyday users' timely and practical information needs. It derives its tasks from trending topics and user comments on popular platforms (e.g., Weibo, Facebook), capturing widely discussed information needs from authentic daily user contexts. DailyReport comprises 150 tasks across two types and 3,546 associated rubrics. These tasks span 10 high-level domains and 35 fine-grained categories, reflecting broad user interests through a multi-level taxonomy. Built on time-sensitive trending topics, DailyReport also supports continuous updates to reflect evolving user needs in real-world scenarios.

We develop a user-centric cascade evaluation pipeline for SAs on these tasks. Consider an authentic user query such as \textit{"List the Chinese universities in the 2026 QS Top 100 rankings, and analyze their respective strengths and weaknesses.”}. If the agent fails to correctly identify the universities, any subsequent analysis becomes meaningless to users. This suggests that rubrics should not be treated independently across dimensions, and different task components have hierarchical priorities from the user perspective. In our pipeline, we decompose each task into subtasks and design cascade rubrics along three disentangled dimensions. We first assess the subtask on the \textbf{instruction following} dimension, and then evaluate \textbf{factuality} and \textbf{rationality} accordingly. Finally, we apply cascade performance attribution to derive interpretable dimensional scores, and further incorporate subtask importance into user-centric performance aggregation to explicitly quantify user preference.

We evaluate \textbf{17} agentic systems across three groups using DailyReport. The results show that existing agents perform well in instruction following, but still struggle with factuality and rationality. Notably, their user preference scores \textbf{remain particularly limited}, revealing a clear gap between current SA outputs and users' perceived expectations. We conduct detailed solving-trace analysis to help diagnose underlying failure patterns and provide valuable guidance for future SA advances.

The structure of DailyReport is shown in Figure \ref{Fig.intro}. In summary, our contributions are as follows:
\begin{itemize}
\item We propose \textbf{DailyReport}, a benchmark for evaluating SAs on daily search tasks. These tasks are grounded in real-world scenarios to reflect authentic user needs. Consisting of 150 tasks and 3,546 rubrics, DailyReport is supported by over 500-hours human annotation.
\item We introduce \textbf{a user-centric cascade evaluation pipeline}. It computes the subtask performance using cascade rubrics along disentangled dimensions, and then enables interpretable dimensional evaluation and explicit user preference quantification accordingly.
\item We conduct \textbf{a thorough empirical assessment} of 17 frontier agentic systems across three groups. The results reveal key strengths and limitations of current search agents.
\end{itemize}

\section{Related Work}

\subsection{Benchmarks for Search Agents}

As SA evolves, several benchmarks have emerged to evaluate their capabilities. The first group \cite{chen2025xbench,li2025mm,song2025bearcubs,wu2026deepresearch} targets fixed-answer tasks that assess information retrieval and multi-step reasoning.
BrowseComp \cite{wei2025browsecomp} serves as a foundational effort evaluating web browsing capabilities.
WideSearch \cite{wong2025widesearch} focuses on wide-context information aggregation requiring the collection of large volumes of atomic facts.
The second group \cite{bigeard2025finance,lyu2025deepshop,huang2026mmdeepresearch} evaluates agents through comprehensive report generation.
DeepResearch Bench \cite{du2025deepresearch} proposes two complementary frameworks assessing report quality and retrieval ability, respectively.
DeepResearch Bench II \cite{li2026deepresearch} collects expert-written investigative reports from reputable open-access venues and constructs research-style tasks following a similar domain distribution.

LiveResearchBench \cite{wang2025liveresearchbench} attempts to align tasks with daily user demands, but remains largely U.S.-centric with limited regional coverage.
As shown in Table \ref{tab.related}, compared with prior works, our benchmark adopts up-to-date daily search tasks that are aligned with real-world user demands. It employs cascade rubrics along disentangled dimensions, enabling interpretable performance attribution and user preference quantification for SA evaluation.

\begin{table*}[h]
    \centering
    
    \resizebox{0.99\textwidth}{!}{%
        \begin{tabular}{l c c c c c c}
        \toprule
            \textbf{Method} & 
            \makecell[c]{\textbf{Open-Ended} \\ \textbf{Task Formats}} & 
            \makecell[c]{\textbf{Daily} \\ \textbf{User Demands}} & 
            \makecell[c]{\textbf{Up-to-date \& } \\ \textbf{Dynamic Evolving}} & 
            \makecell[c]{\textbf{Disentangled} \\ \textbf{Eval Dimension}} & 
            \makecell[c]{\textbf{Cascade} \\ \textbf{Eval Rubrics}} &
            \makecell[c]{\textbf{Quantify} \\ \textbf{User Preference}} \\ 
            \midrule
            BrowseComp         & \xmark & \xmark & \xmark & \cmark & \xmark & \xmark\\
            WideSearch  & \xmark & \xmark & \xmark & \cmark &\xmark & \xmark\\
            DeepResearch Bench  & \cmark & \xmark & \xmark & \xmark &\xmark & \xmark\\
            DeepResearch Bench II  & \cmark & \xmark & \xmark & \xmark &\xmark & \xmark\\
            LiveResearchBench  & \cmark & \cmark & \cmark & \xmark & \xmark & \xmark\\
            ResearchRubrics            & \cmark & \xmark & \xmark  & \xmark & \xmark & \xmark\\
            
            \midrule
            DailyReport (Ours)                     & \cmark & \cmark & \cmark & \cmark & \cmark & \cmark\\
        \bottomrule
        \end{tabular}%
    }
\caption{Comparison of representative benchmarks across task-oriented dimensions (first three columns) and evaluation-oriented dimensions (last three columns).}
\label{tab.related}
\end{table*}

\subsection{Search Agents}
The remarkable progress of LLMs has accelerated the development of SAs \cite{zhou2025memento,xi2025survey}, particularly Deep Research Agents (DRAs) for challenging report-generation tasks. LangChain's Deep Researcher \cite{LearningCircuit2025} performs multi-step web search and synthesizes information locally for response generation. DeepResearcher \cite{zheng2025deepresearcher} scales reinforcement learning with authentic web search interactions for agent training. Tongyi DeepResearch \cite{team2025tongyi} combines agentic mid-training and post-training, enabling scalable reasoning and information seeking across complex tasks. Meanwhile, recent production-grade agents, including Gemini \cite{google2026geminidr}, Grok \cite{xai2026grok3} , and Qwen Deep Research \cite{qwen2026qwendr}, have shown the capability to perform multi-step web exploration and synthesize comprehensive research reports.
Based on these works, DailyReport systematically analyzes the capabilities and limitations of current SAs to further advance this field.

\section{DailyReport Benchmark}

\begin{figure*}[t] 
\centering 
\includegraphics[height=0.52\textwidth,width=0.98\textwidth]{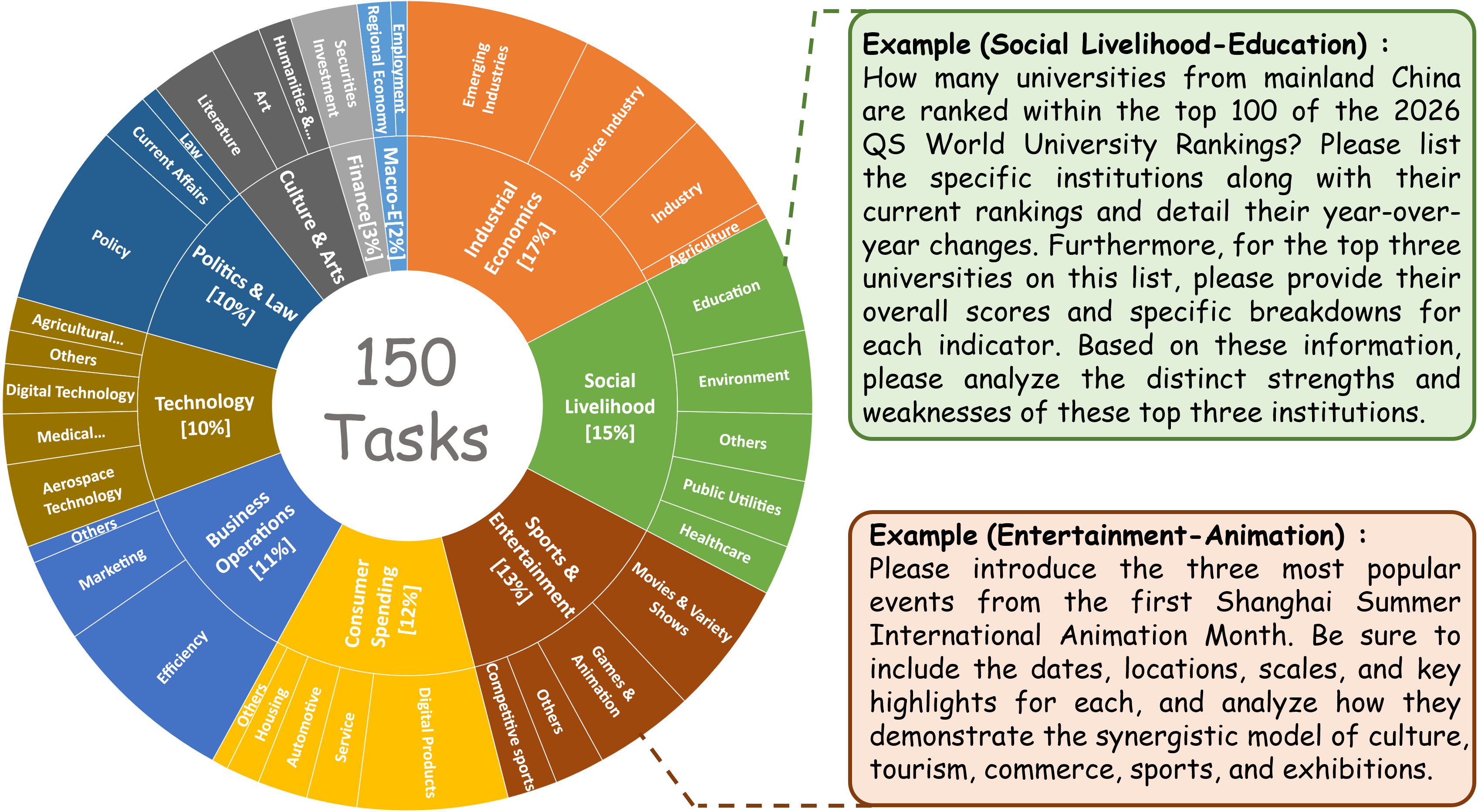} 
\caption{Detailed characteristics of daily search tasks in DailyReport. The benchmark comprises 150 expert-curated tasks with 3,546 detailed rubrics across 10 high-level domains and 35 fine-grained categories. It evaluates search agents in daily user scenarios and aligns closely with predominant real-world user demands.}
\label{Fig.1} 
\end{figure*}

\subsection{Task Characteristic}
Figure \ref{Fig.1} provides the detailed task characteristics of DailyReport. Compared with existing researches, it has the following distinctive features:

\textbf{Tasks are rooted in real-world scenarios and better capture users' daily search needs.} For example, the search task on QS rankings in Figure \ref{Fig.1} is derived from authentic trending topics during the admissions season. It directly reflects practical user interests in university selection and academic planning. In addition, these tasks are framed as broad queries that cover multiple related sub-questions for report generation, which better aligns with how typical users search their needs in real worlds.

\textbf{Tasks are grounded in up-to-date trending topics and continuously evolving.} As illustrated in Figure \ref{Fig.1}, these tasks are consistently grounded in recent real-world events and are regularly updated. This requires agents to iteratively search for information on user-relevant trending topics, rather than relying solely on an LLM’s internal knowledge.

\subsection{Task Construction}
The task construction procedure is primarily conducted by recruited human experts in three stages: \textbf{(1) Trending Topic Selection.}  \textbf{(2) Expert-crafted Task Formulation.} \textbf{(3) Hybrid Topic Annotation.} 

\paragraph{Trending Topic Collection} To root our tasks in real-world scenarios, we primarily select the trending topics from major Western platforms (e.g., Facebook, Reddit, and Twitter) and Chinese platforms (e.g., Weibo, Xiaohongshu, and Zhihu). The collected topic information consists of trending event posts and corresponding user comments, ensuring diverse and regionally representative coverage of authentic user information demands. 

\paragraph{Expert-crafted Task Formulation} 
We recruit human experts to formulate daily search tasks from each topic report and its user comments.
This process yields 150 open-ended tasks that evaluates whether agents can reliably satisfy real users' timely and practical information needs. We set the following requirements for task formulation:
\begin{itemize}
    \item \textbf{Principle:} (1) \textit{Authenticity}: Tasks must be realistic and reflect the genuine information needs of specific user demographics.  (2) \textit{Clarity}: Task descriptions strictly avoid ambiguous phrasing to ensure precise instructions. (3) \textit{Safety}: Tasks are benign to prevent being rejected by the safety mechanisms. 
    \item \textbf{Type:} (1) \textit{100 retrieval-centric tasks}, which focus on retrieving and integrating objective information about specified entities, with only lightweight analysis.  (2) \textit{50 analysis-centric tasks}, which focus on broader subjective topics and require SAs to autonomously identify relevant information for deeper analysis.

\end{itemize}

\paragraph{Hybrid Task Annotation} 
Considering the diversity of daily domains, annotators conduct hybrid task annotation. They first classify each task into 35 fine-grained categories and then consolidate these categories into 10 high-level domains. Fine-grained categories represent specific user interests (like \textit{education}), while high-level domains represent broader fields (like \textit{Social Livelihood}).

\subsection{Rubric Generation}

We decompose each task into subtasks and generate cascade rubrics for each subtask across disentangled dimensions. This process combines LLM-based generation with extensive human refinement, while also supporting full LLM-based automation.

\paragraph{Subtask Decomposition} 

We further categorize commonly emphasized constraints by users into several groups. For instance, \textit{Scope Constraints} dictates that the response must adhere to the boundaries defined in the requirement (e.g., temporal, spatial). Detailed definitions are provided in Appendix \ref{sec:appendixA}.
Our subtasks are then formulated with different combinations of these constraints and generally follow the principles: (1) \textit{Atomicity}: Subtasks must be atomic and a single subtask typically corresponds to one constraint type (excluding \textit{Scope} and \textit{Completeness} constraints, which cannot exist independently). 
(2) \textit{Coverage}: Subtask aggregation must cover every requirement of the original task. (3) \textit{Traceability}: Subtasks must be strictly grounded in the original task to avoid hallucination.

\paragraph{Rubric Formulation}
As shown in Table \ref{tabdim}, we define three disentangled dimensions to evaluate SAs on our daily search tasks. LLMs then formulate cascade rubrics across the three dimensions with human expert assistance for subtask assessment.
Compared with traditional macro-level rubrics, our cascade rubrics support more interpretable performance attribution and allow subtask importance to be incorporated during aggregation.

\begin{table}[H]
    \centering
    \small
    \renewcommand{\tabularxcolumn}[1]{m{#1}} 

    \begin{tabularx}{\linewidth}{>{\centering\arraybackslash}m{1.8cm} X} 
        \toprule
        \textbf{Dimensions} & \multicolumn{1}{c}{\textbf{Description}} \\ 
        \midrule
        
        Instruction Following  & 
        Evaluates the agent's ability to \textbf{accurately} understand and \textbf{fully} execute user instructions. \textbf{(Objective)}\\
        \addlinespace
        
        Factuality  & 
        Evaluates the agent's ability to generate \textbf{factually accurate} content. The verification process requires external search tools. \textbf{(Objective)}\\
        \addlinespace
        
        Rationality  & 
        Evaluates the ability to produce \textbf{logically coherent} reasoning and analysis. The process can be judged solely by cross-referencing the context. \textbf{(Subjective)}\\
        
        \bottomrule
    \end{tabularx}
    \caption{Three disentangled evaluation dimensions which are designed to be strictly orthogonal.}
    \label{tabdim}
\end{table}

\section{User-centric Cascade Evaluation}

\subsection{Rubric Assessment}
We employ cascade rubrics across the three dimensions for subtask evaluation. Let $T_i$ denote the \textit{i}-th subtask, and $\mathrm{Res} = \mathrm{SA}(T_i)$ denote the agent’s response to $T_i$, where $i \in [1, n]$ and $n$ is the total number of subtasks. The dimensional score $\mathrm{dim}_i$ is calculated as in Eq.\ref{eq0}, where $\mathrm{dim} \in \{\mathrm{ins}, \mathrm{fac}, \mathrm{rat}\}$. For each dimension, the judge model $\operatorname{Judge}_{\mathrm{dim}}$ evaluates $\mathrm{Res}$ against the corresponding rubric $r_{\mathrm{dim}}(T_i)$. More judgment details are provided in Appendix \ref{sec:appendixB}. This produces three subtask scores, $\mathrm{ins}_i$, $\mathrm{fac}_i$, and $\mathrm{rat}_i$, which represent the dimensional performance of the \textit{i}-th subtask.

\begin{equation}
    \mathrm{dim}_i = \operatorname{Judge}_{\mathrm{dim}} \Big( T_i, \mathrm{Res}, r_{\mathrm{dim}}(T_i) \Big)
\label{eq0}
\end{equation}

\subsection{Cascade Performance Attribution}
We derive an interpretable overall score for each dimension by accounting for subtask performance dependencies among the three dimensions.

\paragraph{Instruction Following}
For a given SA system, we directly obtain its subtask performance $\mathrm{ins}_i$ on instruction following dimension using Eq.\ref{eq0}. The score $\mathrm{ins}_i \in \{0,0.5,1\}$ reflects whether $\mathrm{Res}$ fully, partially, or fails to satisfy the corresponding rubric. The overall score $\mathrm{Ins}$ is defined as the average performance across all subtasks, as shown in Eq.\ref{eq2}. 

\begin{equation}
    \text{Ins} = \frac{\sum_{k=1}^{n} \mathrm{ins}_k}{n}
\label{eq2}
\end{equation}

\begin{algorithm}[t]
\caption{User-Centric Aggregation}
\label{alg1}
\begin{algorithmic}[1]
    \Require $p_k \in \{P0, P1, P2(a), P2\}$, scores $o_k \in [0,1]$
    
    \State $\mathcal{S}_0 \gets \{k : p_k = P0\}, \mathcal{S}_1 \gets \{k : p_k = P1\}$
    \State $\mathcal{S}_2^{(a)} \gets \{k : p_k = P2(a)\}$ for each group $a \in \{1, \dots, N\}$
    
    \State $\mathcal{G} \gets  \text{Mean}\{o_k : k \in \mathcal{S}_2^{(a)}\}$
    \State $c_0 \gets \text{Mean}\{o_k : k \in \mathcal{S}_0\}$ \textbf{ if } $\mathcal{S}_0 \neq \emptyset$ \textbf{ else } $1$
    \State $c_1 \gets \text{Mean}\{o_k : k \in \mathcal{S}_1\} \cup \mathcal{G}$
    
    \If{$\forall\, k: o_k = 1$} \Return \textbf{UserPref} $= 4$
    \EndIf 
    
    \If{$c_0 = 0 \lor c_1 < 0.3 \lor (c_0 < 0.5 \land c_1 < 0.5)$} \Return \textbf{UserPref} $= 1$ 
    \EndIf
    
    \State $v_1 \gets \forall\, k \in \mathcal{S}_1 : o_k > 0$
    \State $v_2 \gets \exists\, k \in \bigcup_a \mathcal{S}_2^{(a)} : o_k > 0$
    
    \If{$c_0 \geq 0.5 \land v_1 \land v_2 \land c_1 \geq 0.7$}
        \State \Return \textbf{UserPref} $= 3$ 
    \Else
        \State \Return \textbf{UserPref} $= 2$ 
    \EndIf
\end{algorithmic}
\end{algorithm}

\paragraph{Factuality} 
We perform cascade performance attribution to obtain reliable factuality performance, where the factuality dimension is considered only if the response satisfies the corresponding instruction following requirement. Otherwise, the required target content is absent, and evaluating its factuality is no longer meaningful. Inspired by this, we define the overall factuality score as Eq.\ref{eq3}.
\begin{equation}
    \mathrm{Fac} = \frac{\sum_{k=1}^{n} \delta_k \cdot \mathrm{ins}_k \cdot \mathrm{fac}_k}{\sum_{k=1}^{n} \delta_k \cdot \mathrm{ins}_k},
\label{eq3}
\end{equation}
where $\delta_k = 1$ if the subtask includes the factuality rubric, and $\delta_k = 0$ otherwise. We first extract the objective claims in $\mathrm{Res}$, along with their supporting references if available. The judge model then verifies each claim using web search and assigns the factuality score $\mathrm{fac}_k \in [0,1]$ accordingly.

\paragraph{Rationality} Similarly, we formulate the overall rationality score in Eq.\ref{eq4}, where $\varphi_k$ indicates whether the \textit{k}-th subtask contains the rationality rubric. The score $\mathrm{rat}_k \in \{0,0.5,1\}$ is assigned by the judge model based on whether $\mathrm{Res}$ is logically reasonable. To reduce its coupling with factuality, the judge model primarily focuses on the subjective reasoning and analytical part of $\mathrm{Res}$.
\begin{equation}
    \mathrm{Rat} = \frac{\sum_{k=1}^{n} \varphi _k \cdot \mathrm{ins}_k \cdot \mathrm{rat}_k}{\sum_{k=1}^{n} \varphi_k \cdot \mathrm{ins}_k}
\label{eq4}
\end{equation}

\subsection{User-centric Performance Aggregation}

We apply user-centric aggregation for subtask performance to obtain the user preference score. First, we define four user preference levels according to real users' perceived helpfulness: \textbf{1 (Unhelpful):} The response entirely misses the user's core needs and is almost unusable for users. \textbf{2 (Deficient):} The response satisfies some user requirements, but contains significant flaws that negatively impact the user experience. \textbf{3 (Acceptable):} The response satisfies the primary user needs, with only minor flaws that do not significantly affect the overall experience. \textbf{4 (Perfect):} The response fully satisfies the user's needs with almost no errors.

Then, we recruit the task creators to conduct an ablation study for each subtask to obtain its user-perceived importance. Specifically, they estimate the user preference level when only the target subtask is left unsatisfied, and assign its importance according to the following mapping: \textbf{P0}: if the resulting response is rated as 1 (Unhelpful), \textbf{P1}: if rated as 2 (Deficient), \textbf{P2}: if rated as 3 (Acceptable).
\textbf{P2(a)} denotes subtasks that are 3 (acceptable) when missed alone but can cause 2 (Deficient) when multiple such subtasks are missed.

Moreover, we calculate the overall subtask performance as Eq.\ref{eq5} according to user experience, where factuality and rationality are meaningful only when the response follows the instructions.
\begin{equation}
    o_k =  \frac{1}{2} \cdot \mathrm{ins}_k \cdot (\mathrm{fac}_k+\mathrm{rat}_k)
\label{eq5}
\end{equation}

Finally, the user-centric aggregation algorithm is developed as in Alg.\ref{alg1}, which aggregates subtask performance $o_k$ based on its importance and computes the overall user preference score $\mathrm{UserPref}$. 

\begin{table*}[t]
\vspace{-3em}
\centering
\resizebox{0.98\textwidth}{!}{%
\begin{tabular}{lccccc}
\toprule
\textbf{Model} & \textbf{UserPref} & \textbf{SubTask Pass} & \textbf{InstFollow} & \textbf{Factuality} & \textbf{Rationality} \\
\midrule
\multicolumn{6}{c}{\textit{Deep Research Agents}} \\
\midrule
OpenAI o3 Deep Research \cite{openai2025o3deepresearch}       
& 2.42 & 0.228 & 0.967 & 0.616 & 0.856 \\
OpenAI o4-mini Deep Research \cite{openai2026o4minidr} 
& 2.40 & 0.241 & 0.961 & 0.663 & 0.778 \\
Gemini Deep Research \cite{google2026geminidr}          
& 2.41 & 0.184 & 0.973 & 0.635 & 0.765 \\
Qwen Deep Research \cite{qwen2026qwendr}             
& 2.17 & 0.119 & 0.934 & 0.612 & 0.662 \\
Grok 3 Deep Research \cite{xai2026grok3}             
& 2.48 & 0.301 & 0.917 & 0.731 & 0.909 \\
\midrule
\multicolumn{6}{c}{\textit{LLMs with Search Tools}} \\
\midrule
Claude Opus 4.6 \cite{anthropic2026claude46}           
& 2.79 & 0.261 & 0.976 & 0.796 & 0.820 \\
GPT 5.4 \cite{openai2026gpt54}                      
& \textbf{2.89} & \textbf{0.484} & \uline{0.982} & \textbf{0.835} & \uline{0.930} \\
Gemini 3.1 Pro \cite{google2026gemini31pro}              
& 2.63 & 0.291 & 0.976 & 0.730 & 0.802 \\
GLM 5 \cite{zhipu2026glm5}                     
& 2.68 & 0.250 & 0.972 & 0.784 & 0.775 \\
Kimi K2.5 \cite{moonshot2026k25}                  
& 2.60 & 0.215 & 0.970 & 0.728 & 0.786 \\
Qwen 3.5 \cite{qwen2026qwen35}                    
& 2.67 & 0.208 & 0.960 & 0.776 & 0.757 \\

\midrule
\multicolumn{6}{c}{\textit{LLMs with Claude Code}} \\
\midrule
CC-Claude Opus 4.6 \cite{anthropic2026claude46}           
& 2.65 & 0.206 & 0.971 & 0.756 & 0.809 \\
CC-GPT 5.4 \cite{openai2026gpt54}                      
& \uline{2.87} & \uline{0.478} & \textbf{0.989} & \uline{0.813} & \textbf{0.933} \\
CC-Gemini 3.1 Pro \cite{google2026gemini31pro}              
& 2.58 & 0.262 & 0.971 & 0.684 & 0.821 \\
CC-GLM 5 \cite{zhipu2026glm5}                     
& 2.65 & 0.265& 0.965 & 0.767 & 0.809 \\
CC-Kimi K2.5 \cite{moonshot2026k25}                      
& 2.61 & 0.223 & 0.964 & 0.718 & 0.796 \\
CC-Qwen 3.5 \cite{qwen2026qwen35}                       
& 2.51 & 0.199 & 0.967 & 0.718 & 0.782 \\
\bottomrule
\end{tabular}}
\caption{Evaluation results of 17 system settings on DailyReport across three categories. \textbf{Bold values} indicate the highest score in each column, while \uline{underlined} denotes the second highest.}
\label{tab:main}
\end{table*}

\section{Experiment}

\subsection{Experiment Setup}

We conduct a comprehensive evaluation of 17 agentic systems in three groups: native DRAs, search-augmented LLMs, and LLMs with Claude Code. We select Gemini-3-flash as the judge model and enabled reasoning mode for all evaluated models.

\subsection{Main Results}
Table \ref{tab:main} illustrates the main results of frontier agentic systems on DailyReport. Overall, LLMs with search tools achieve the best performance, followed by LLM equipped with Claude Code, while native DRAs obtain relatively lower scores. Among all systems, GPT 5.4-based configurations performs best.
This suggests that daily search tasks benefit from the combination of direct web search and strong general-purpose LLMs. In contrast, Claude Code is optimized for code-oriented workflows, which may introduce redundant context and lead to suboptimal results on search-intensive tasks. Native DRAs may rely on specialized internal models to balance cost, latency, and stability, making them less effective than stronger general models.

\textbf{Current systems are particularly weak on UserPref}: even the highest score remains below the acceptable level of 3, showing that they still struggle to produce consistently satisfactory responses. For further comparison, we report SubTask Pass, the proportion of subtasks satisfying all rubric criteria, which remains low across systems. A system may achieve higher UserPref despite lower SubTask Pass when it satisfies more high-importance subtasks while missing less critical ones.

\begin{figure*}[t]
	\centering
	\subfigure{\includegraphics[width=.325\textwidth,height=.262\textwidth]{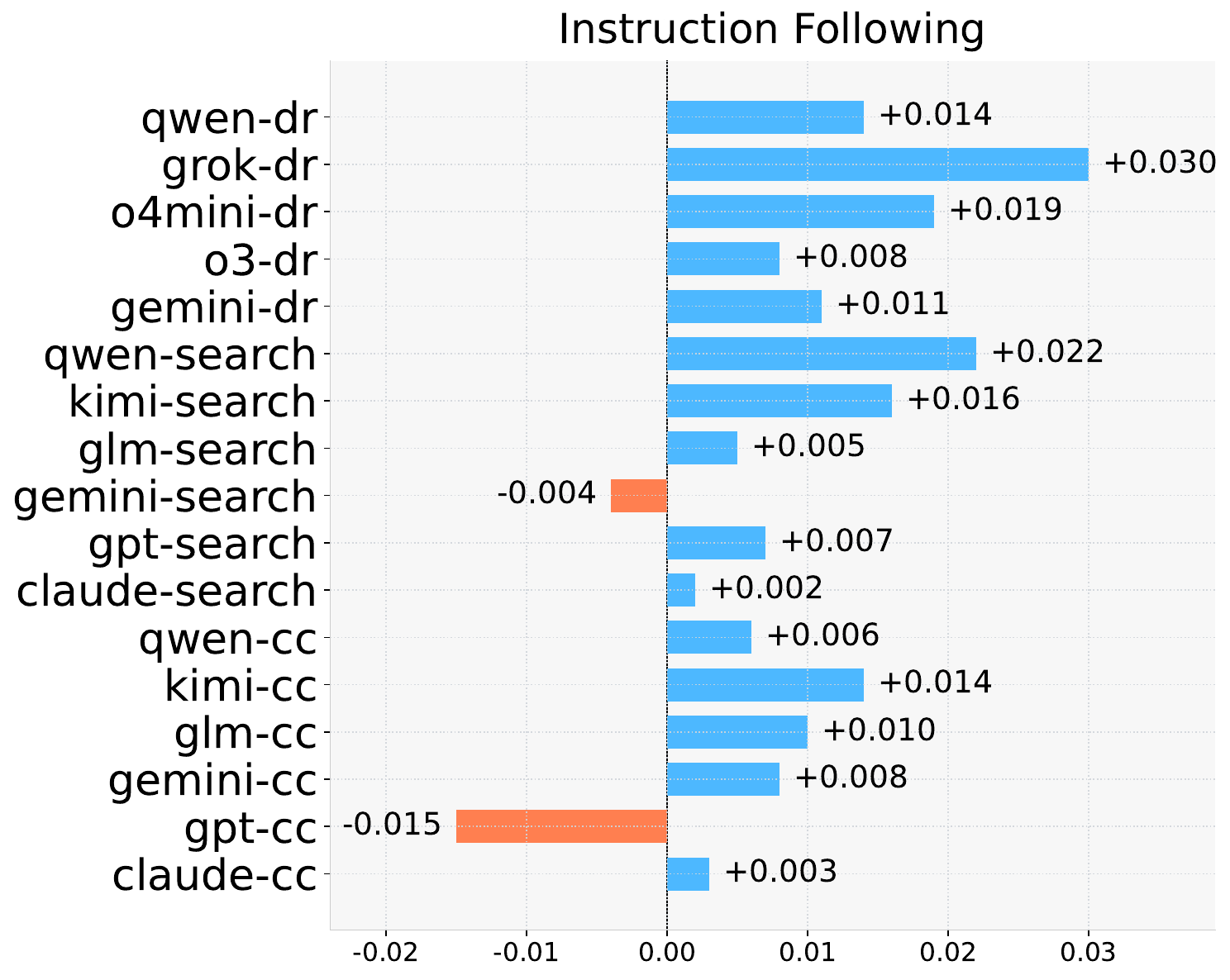}}
	\subfigure{\includegraphics[width=.325\textwidth]{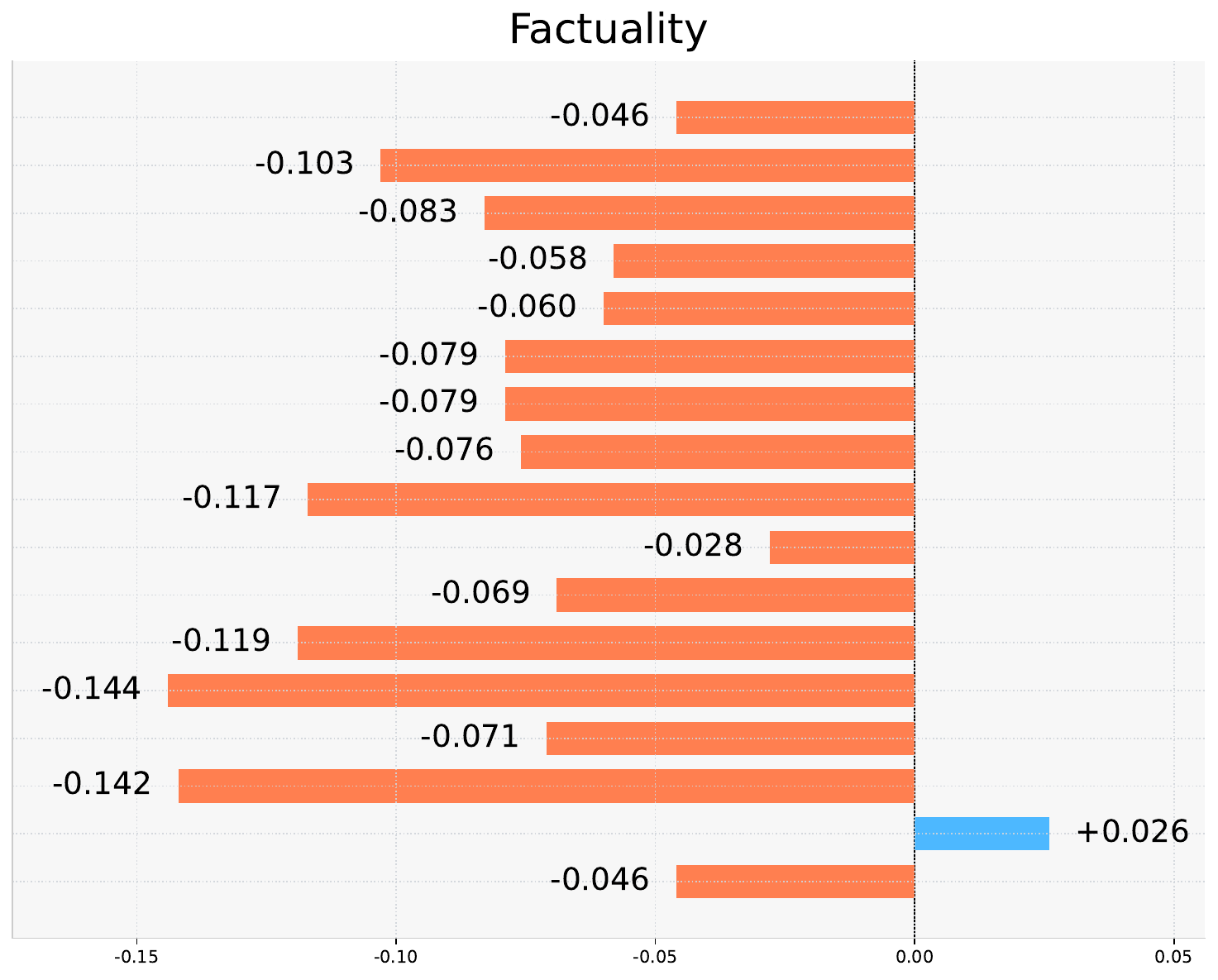}}
        \subfigure{\includegraphics[width=.325\textwidth,height=.262\textwidth]{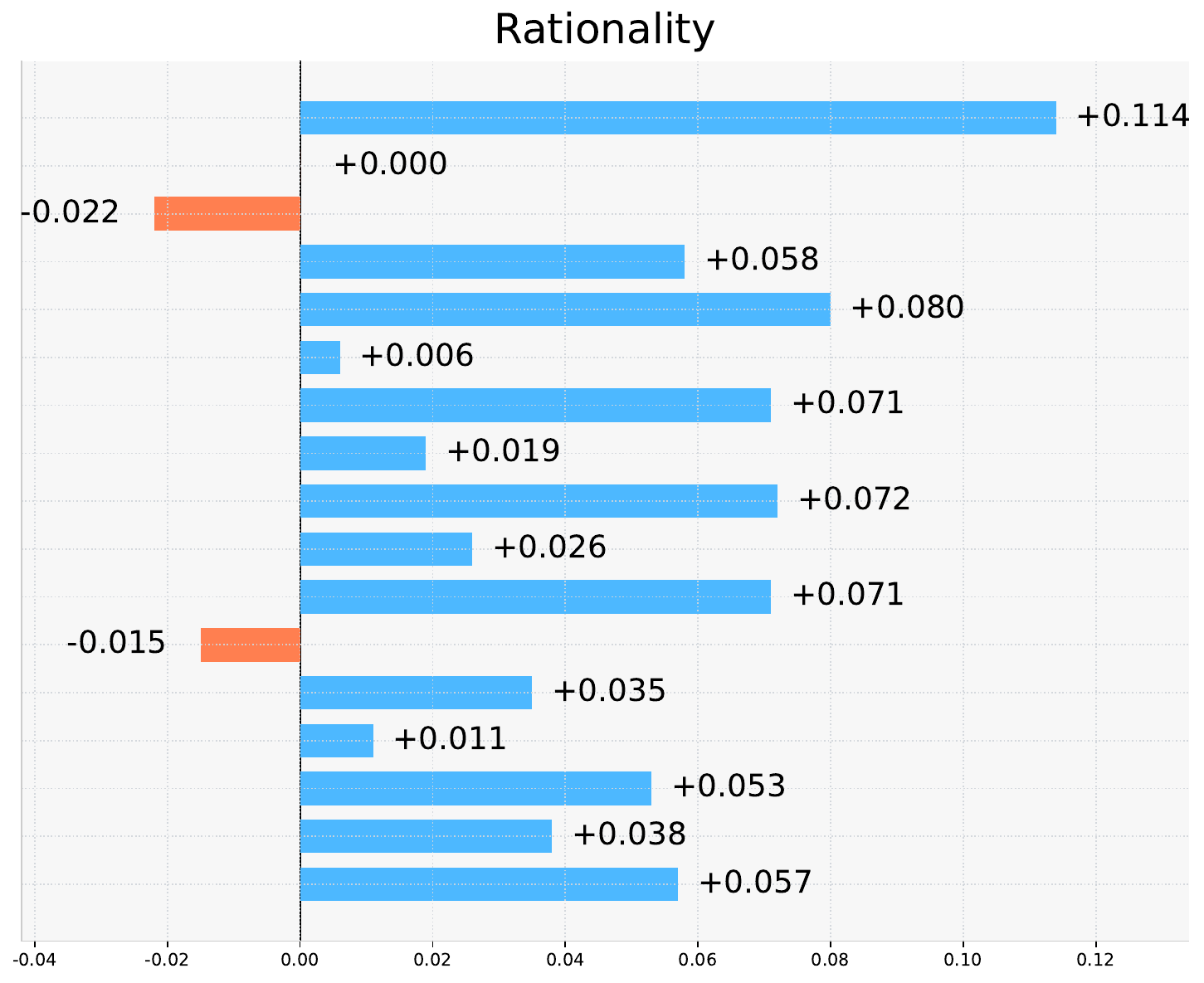}}
	\caption{Task type effect across three dimensions. For each model, we report the difference $\Delta = \mathrm{Avg}_{\mathrm{analysis}} - \mathrm{Avg}_{\mathrm{retrieval}}$ between its average scores on 50 analysis-centric and 100 retrieval-centric tasks. Blue bars indicate $\Delta>0$ and stronger analysis-centric task performance, while yellow bars indicate the opposite.}
	\label{Fig.type}
\end{figure*}

\begin{figure*}[t]
	\centering
	\subfigure{\includegraphics[width=.48\textwidth]{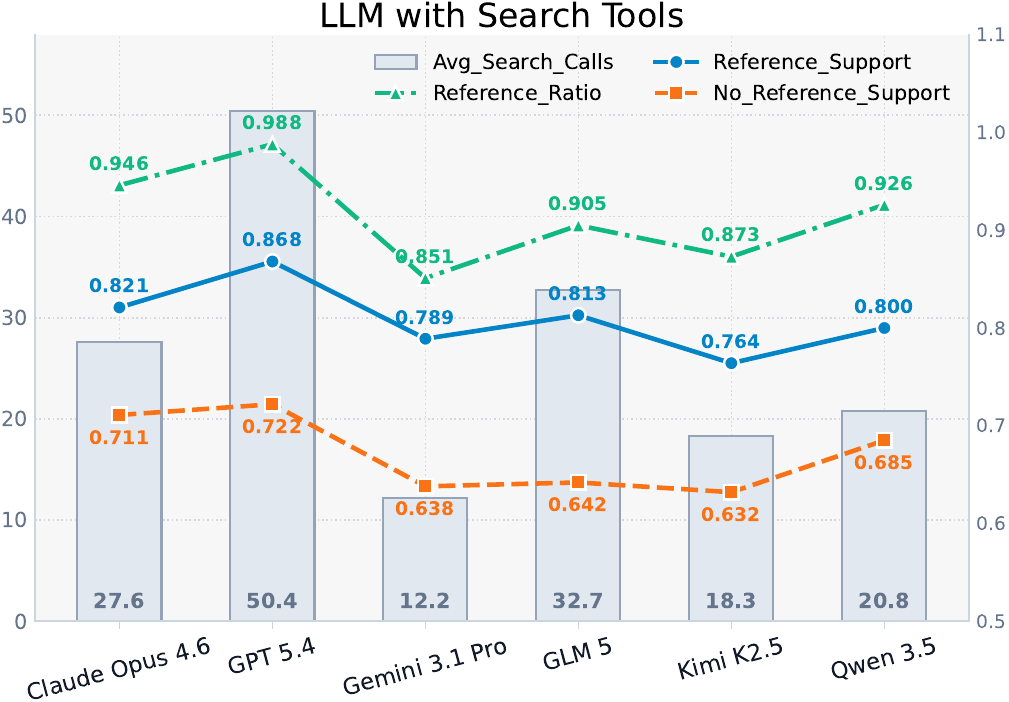}}
	\subfigure{\includegraphics[width=.48\textwidth]{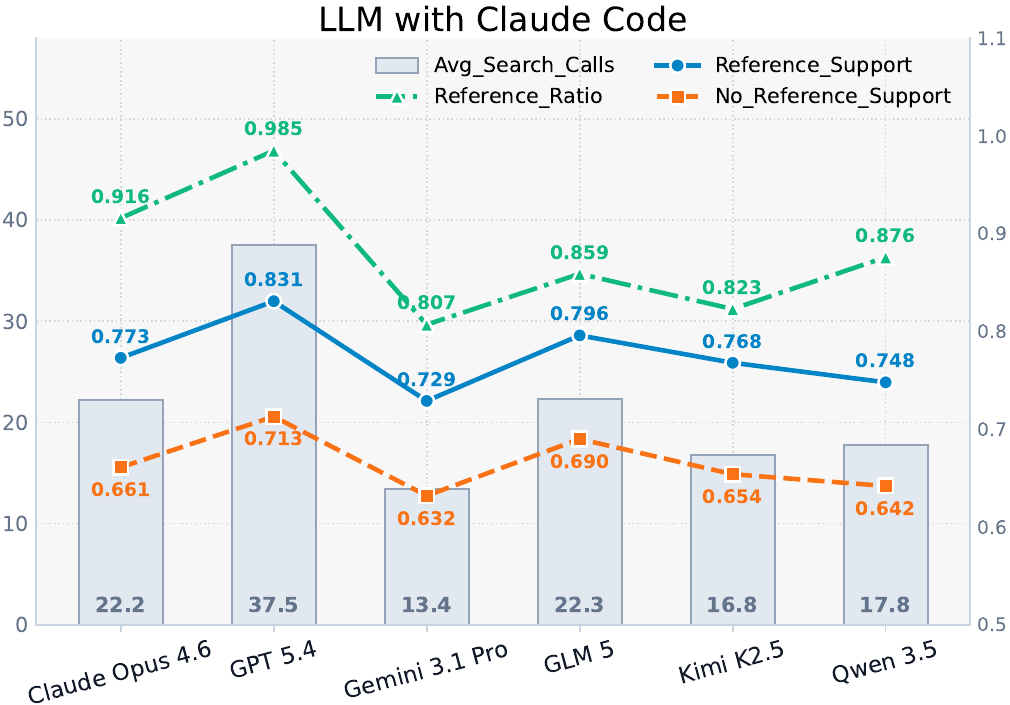}}
\caption{Trace Analysis. Avg\_Search\_Calls measures the total number of search-tool calls. Reference\_Ratio measures the proportion of claims that are supported by references, Reference\_Support measures the factual accuracy of claims with references, and No\_Reference\_Support measures the factual accuracy of claims without references.}
\label{Fig.tra}
\end{figure*}

The dimensional scores reveal different capability bottlenecks of current agentic systems. First, all systems achieve relatively high InstFollow scores, suggesting that frontier models generally possess strong instruction-following abilities and can cover most explicit user requirements. In contrast, Factuality remains the weakest dimension, indicating that systems still struggle to acquire accurate and timely evidence to avoid hallucinated claims. Rationality is still far from perfect, possibly because reasoning over trending topics often involves incomplete, timely, or conflicting information.

\subsection{Task Type Analysis}
Task type effects across three dimensions are shown in Figure \ref{Fig.type}. In total, analysis-centric tasks show slightly better instruction following and rationality, but lower factuality. Specifically, analysis-centric tasks are more open-ended and usually provide broader analytical requirements, making it easier for models to cover the requested aspects and obtain higher InstFollow scores. However, open-ended analysis also leads to more divergent search paths. The retrieved evidence is often scattered across heterogeneous sources, so claims are harder to triangulate through cross-source verification than in retrieval-centric tasks. As a result, models are more likely to introduce unsupported factual claims and suffer from lower factuality. The stronger rationality performance on analysis-centric tasks can be explained by their focus on topic-level summarization and subjective analysis, which better match models' strengths in open-ended analytical writing. In addition, the task formulations usually provide explicit analytical directions that help models organize coherent explanations and arguments.

\subsection{Trace Analysis}
We analyze the solving traces of each system in Figure \ref{Fig.tra}. 
Search-tool usage directly reflects the extent of retrieval and iterative reasoning, which shows the strongest association with overall performance. This suggests that \textbf{future SAs should incorporate mechanisms to ensure sufficient retrieval before generation.} Compared to search-augmented LLMs, LLMs with Claude Code invoke search tools less frequently, possibly because the code-oriented framework encourages context reuse and avoids unnecessary tool calls for efficiency. 

We additionally examine the weaker factuality dimension through three reference-related metrics. Most systems achieve a high Reference\_Ratio, indicating that they tend to support generated claims with references. This improves the factual accuracy over unsupported claims to some extent. However, Reference\_Support remains lower, showing that citing references does not always guarantee factual correctness. This highlights that \textbf{future SAs need to improve reference quality and reference-claim alignment}, which are still inadequate in current systems, as further analyzed in Appendix \ref{sec:appendixB}.

\subsection{Meta Evaluation}

\begin{wraptable}{l}{0.46\textwidth}
\centering
\vspace{-12pt} 
\setlength{\tabcolsep}{3pt} 
\resizebox{\linewidth}{!}{
        \begin{tabular}{lcccc}
        \toprule
        Models  &UserP &Ins (\%) &Fac (\%) &Rat (\%) \\
        \midrule
        GPT-search &$2.90_{\scriptscriptstyle \pm 0.007}$ & $98.3_{\scriptscriptstyle \pm 0.3}$ &$83.6_{\scriptscriptstyle \pm 0.2}$ &$93.4_{\scriptscriptstyle \pm 0.4}$ \\
        Claude-search  &$2.78_{\scriptscriptstyle \pm 0.010}$ &$97.8_{\scriptscriptstyle \pm 0.2}$ & $78.5_{\scriptscriptstyle \pm 1.0}$ &$81.5_{\scriptscriptstyle \pm 0.5}$ \\
        Gemini-search &$2.64_{\scriptscriptstyle \pm 0.010}$ &$97.7_{\scriptscriptstyle \pm 0.1}$ &$69.9_{\scriptscriptstyle \pm 2.7}$ &$80.5_{\scriptscriptstyle \pm 0.9}$ \\
        \bottomrule
\end{tabular}}
\caption{Robustness Analyses.}
\label{tab:meta_robu}
\vspace{-10pt} 
\end{wraptable}

\paragraph{Robustness}
Evaluation stability reflects the reproducibility and practical usability of a benchmark, yet it is often ignored in existing open-ended SA benchmarks. We conduct a robustness analysis on DailyReport by selecting three representative models and repeating the evaluation three times. We use the standard deviation across runs to measure evaluation stability. As shown in Table \ref{tab:meta_robu}, the results exhibit low variance, demonstrating that DailyReport provides stable results. This supports its practical value as a reliable benchmark for SAs.

\paragraph{Judge Model Selection}

\begin{wraptable}{r}{0.49\textwidth}
\centering
\vspace{-12pt} 
\setlength{\tabcolsep}{2pt} 
\resizebox{\linewidth}{!}{
        \begin{tabular}{lccccc}
        \toprule
        Models  &Ins (\%)  &Fac (\%) &Rea (\%) &Avg.Cost (\$)\\
        \midrule
        GPT-5.2 & 92.1 &91.7 &91.4 & 2.04 \\
        Gemini-2.5-Pro  & 94.5 & 93.1 &93.8 & 1.58 \\
        Claude-4.5-Sonnet & 95.1 & 94.5 &95.7 & 2.53 \\
        \midrule
        Gemini-3-flash & 96.5 & 94.2 &95.3 &0.45 \\
        \bottomrule
\end{tabular}}
\caption{Accuracy and cost of different judge LLMs.}
\label{tab:meta_eval}
\vspace{-10pt} 
\end{wraptable}
We conduct a meta-evaluation to compare different LLMs as evaluators. Each LLM evaluates the same set of reports, and we compute its accuracy against human expert annotations, as reported in Table \ref{tab:meta_eval}. Gemini-3-Flash follows our criteria more accurately than GPT-5.2 and Gemini-2.5-Pro, while achieving comparable agreement to Claude 4.5 Sonnet. Considering both evaluation accuracy and cost, we select Gemini-3-Flash as the judge model for all experiments.



\paragraph{Metric Validation}
To validate our metrics, we conduct a meta-evaluation on 300 randomly sampled subtasks. For instruction-following, human annotators and the judge model independently evaluate these samples. Final labels are determined through adjudication, where experts review both results to make more informed decisions that approximate the ground truth.
Our evaluation achieves 96.5\% accuracy, substantially exceeding human annotation accuracy of 88.4\%. Since factuality and rationality are difficult for humans to annotate in long reports, we instead assess these metrics through manual spot checks. The accuracy reaches 94.2\% for factuality and 95.3\% for rationality, meeting the expected requirements. For user preference, users are given the generated reports and subtask results, and assign an overall score from 1 to 4 to indicate their preference. The agreement heatmap is in Appendix \ref{sec:appendixB}. UserPref achieves high agreement with real user ratings, with a Weighted Cohen's Kappa score of 0.859. This suggests that it effectively reflects real users' perceived experience.

\begin{figure*}[t] 
\centering 
\includegraphics[width=0.99\textwidth]{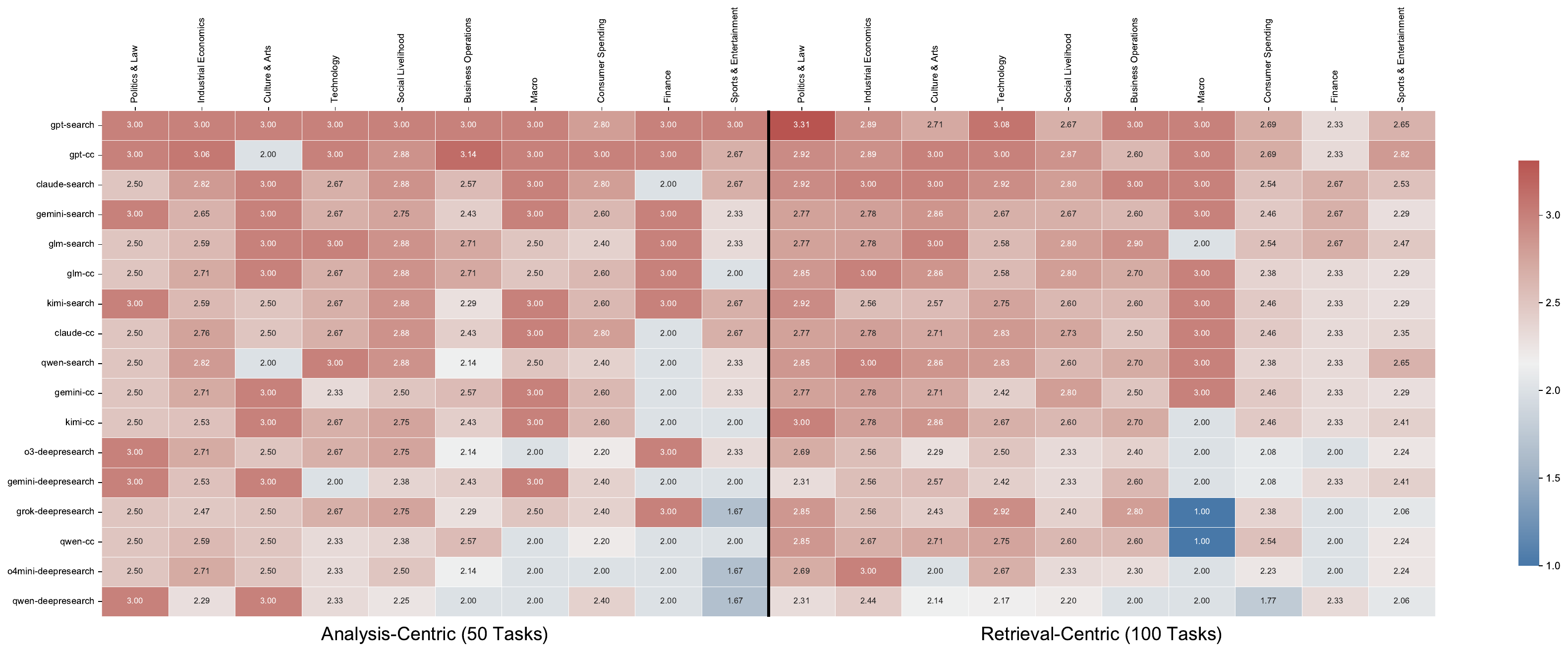} 
\caption{Domain distribution. The heatmap reports the average UserPref scores of different systems on analysis-centric and retrieval-centric tasks across 10 high-level domains.}
\label{Fig.topic} 
\end{figure*}

\subsection{Domain Distribution}
UserPref across 10 high-level domains is shown in Figure \ref{Fig.topic}. Systems generally achieve higher user preference in domains such as Politics \& Law and Industrial Economies, where information is more structured and can be verified through authoritative sources, such as official announcements, institutional reports, or mainstream news coverage. In contrast, domains such as Sports \& Entertainment tend to receive lower scores, as they often involve rapidly changing events and subjective user opinions, making it harder to retrieve comprehensive evidence and produce reliable analysis. This domain-level variation suggests that current search agents perform better on topics with stable and well-documented evidence, but still struggle with highly dynamic or subjective information needs.

\section{Conclusion}
In this work, we present an open-ended benchmark (DailyReport) to evaluate search agents on daily search tasks. It contains 150 tasks with 3,546 associated rubrics, capturing widely discussed and timely information needs of real-world users. We decompose each task into subtasks and design cascade rubrics along disentangled dimensions for subtask evaluation. Through cascade performance attribution and user-centric aggregation, DailyReport produces interpretable dimensional scores and an additional user preference score. Finally, we conduct an empirical assessment of 17 agentic systems to characterize current search agents and offer insights for future research in this area.

\section*{Acknowledgments}
We would like to thank Ruyu Ruan, Yinglong Deng, Yi Shi, Jianfei Zhao, Jiayi Guo, Hao Zheng, Zhiqiang Li, Mingyue Yuan, Danni Li, Ting Zeng, Xin Tang, Luju Gao, Zixi Yuan, and Tingting Liang for their valuable contributions to the benchmark construction process.

\bibliographystyle{plainnat}
\bibliography{custom}

\newpage
\appendix

\section{Construction Appendix}
\label{sec:appendixA}
\subsection{Human Annotation}
DailyReport involved substantial human participation across the entire task construction pipeline. This process involved over 500 hours of human annotation and review, with all contributors compensated at approximately USD 56–70 per day for their work. We recruited contributors with diverse backgrounds, including different regions, educational experiences, online platform habits, and domain familiarity. All contributors were familiar with both Western and Chinese media ecosystems, enabling them to better identify users' daily information needs from trending topic contexts and user comments. Before annotation, they were given detailed guidelines on authenticity, clarity, safety, tool dependency, unimodality, and disentanglement, and completed pilot examples to ensure a consistent understanding of the construction criteria. 

During task construction, annotators first reviewed public trending posts and user comments to identify common information needs, while filtering out topics that were unsafe, overly narrow, ambiguous, or not suitable for generating daily reports. Task writers then transformed selected topics into realistic search tasks with clear scopes, factual requirements, and analytical components. Additional reviewers checked each task for clarity, realism, safety, search dependency, and category consistency. The original task creators also participated in estimating subtask importance for user-centric aggregation, helping the final benchmark reflect not only whether a system satisfies individual requirements, but also how much each requirement matters to user experience.

\begin{wrapfigure}{r}{0.49\textwidth}
\centering
\vspace{-15pt} 
\includegraphics[width=0.47\textwidth]{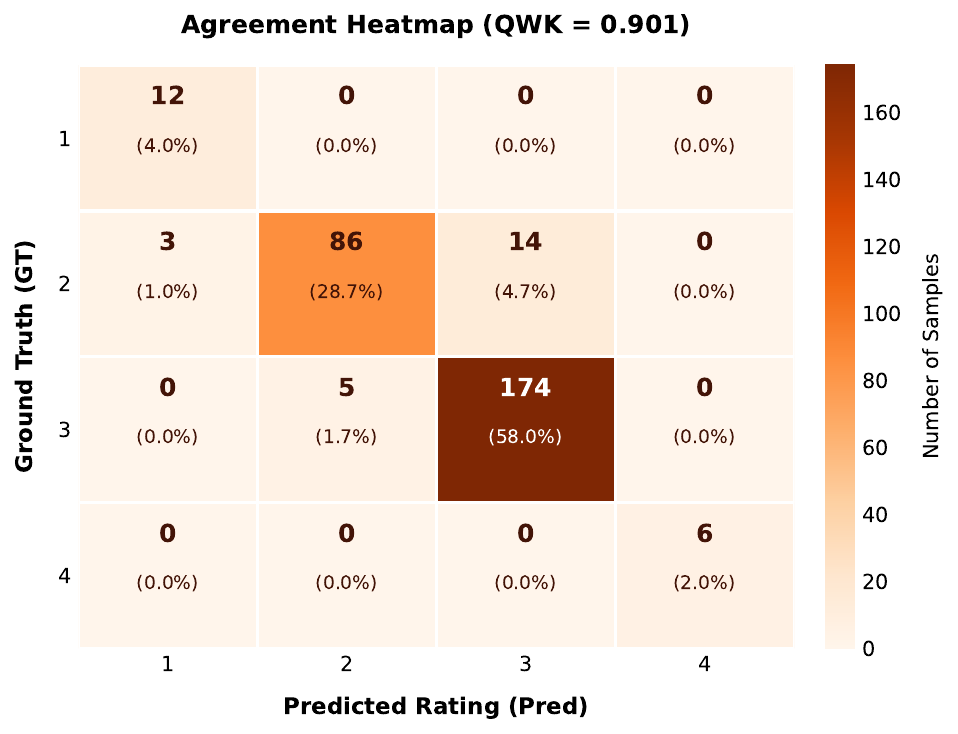} 
\caption{Agreement heatmap. Each cell shows the number of sampled instances with the corresponding score pair, and the diagonal concentration indicates strong consistency with real users' perceived experience.}
\label{Fig.maph}
\vspace{0pt} 
\end{wrapfigure}

\subsection{Constraints Elaboration}
We define the constraint categories as follows, which are utilized to decompose the constructed tasks and derive the corresponding subtasks. 
Specifically, the categories include: 
(1) \textbf{Content Constraints}, which concern the core information elements to be outputted; 
(2) \textbf{Scope Constraints}, which require the generated content to strictly remain within the boundaries specified in the requirement prompt, such as temporal, spatial, domain, source, or policy restrictions; 
(3) \textbf{Completeness Constraints}, which require the output to satisfy specific standards of quantity, exhaustive coverage, and informational completeness; 
(4) \textbf{Quantity Constraints}, which define exact measurable targets for the output, including word counts, item quantities, and overall length; 
(5) \textbf{Format Constraints}, which specify the structural layout, styling, and formatting of the generated response; 
(6) \textbf{Setting Constraints}, which require the agent to operate strictly within the given settings, without violating designated backgrounds, character personas, scenarios, prerequisites, or provided data; 
(7) \textbf{Attribute Constraints}, which specify the stylistic and perspectival properties of the output; 
(8) \textbf{Action \& Rule Constraints}, which define the exact actions, execution paths, methodologies, or logical rules the agent must follow to generate the output; and 
(9) \textbf{Function Constraints}, which require the output to serve a specific practical function, achieve a targeted effect, or solve a defined problem.

\begin{table*}[h]
\centering
\resizebox{0.98\textwidth}{!}{%
\begin{tabular}{lcccc}
\toprule
\textbf{Model} & \textbf{Reference Accuracy} & \textbf{Refer-Claim Consistency} & \textbf{Web Search} & \textbf{Web Content Mining} \\
\midrule
\multicolumn{5}{c}{\textit{Deep Research Agents}} \\
\midrule
OpenAI o3 Deep Research      
& 0.550 & 0.858 & -  & - \\
OpenAI o4-mini Deep Research 
& 0.585 & 0.784 & - & - \\

\midrule
\multicolumn{5}{c}{\textit{LLMs with Search Tools}} \\
\midrule
Claude Opus 4.6            
& 0.739 & 0.845 & 15.7 & 11.9 \\
GPT 5.4                       
& 0.814 & 0.906 & 31.6 & 18.8 \\
Gemini 3.1 Pro              
& 0.705 & 0.824 & 9.4 & 2.8 \\
GLM 5                     
& 0.710 & 0.792 & 17.7 & 15.0 \\
Kimi K2.5                       
& 0.762 & 0.787 & 13.7 & 4.6 \\
Qwen 3.5                        
& 0.654 & 0.769 & 9.6 & 11.2 \\

\midrule
\multicolumn{5}{c}{\textit{LLMs with Claude Code}} \\
\midrule
CC-Claude Opus 4.6          
& 0.701 & 0.850 & 17.1 & 5.1 \\
CC-GPT 5.4                
& 0.741 & 0.897 & 22.5 & 15.0 \\
CC-Gemini 3.1 Pro              
& 0.768 & 0.834 & 11.5 & 1.9 \\
CC-GLM 5                     
& 0.773 & 0.846 & 14.6 & 7.7 \\
CC-Kimi K2.5                     
& 0.772 & 0.807 & 12.7 & 4.1 \\
CC-Qwen 3.5                      
& 0.679 & 0.739 & 11.5 & 6.3 \\
\bottomrule
\end{tabular}}
\caption{Detailed analysis of solving traces. Reference Accuracy evaluates the factual reliability of cited references. Refer-Claim Consistency measures whether generated claims are accurately supported by their cited references. Web Search counts calls to web search tools such as Serper for retrieving search results and snippets, while Web Content Mining counts calls to webpage-fetching tools such as Jina for accessing full webpage content. Both types of calls are treated as important search-tool operations for evidence gathering. Results for some closed-source Deep Research Agents are partially omitted, as they do not expose the internal traces (e.g., search queries, visited URLs) required for reliable measurement of certain metrics.}
\label{tab:search}
\end{table*}

\section{Evaluation Appendix}
\label{sec:appendixB}

\subsection{Meta Evaluation}
To validate the user preference score corresponds to the real users' perceived experience, we conduct a meta-evaluation for the user preference. The generated reports and subtask results of 300 randomly sampled subtasks are provided to diverse users who are asked to assign an overall task score from 1 to 4 to indicate their preference. Figure \ref{Fig.maph} shows that for tasks with real user preferences of 1 and 4, the user preference scores aggregated by our method achieve high alignment, and this high consistency is also maintained in the scores of 2 and 3. The Weighted Cohen’s Kappa score of 0.859 for our meta-evaluation verifies the high degree of alignment between the user preference score and the real user rating. This suggests that the user preference score effectively reflects real users’ preference.


\subsection{Search Analysis}
Overall, the two reference-related metrics remain far from ideal, with Reference Accuracy being particularly limited. This suggests that current search agent systems may still rely on inaccurate, unreliable, or inappropriate citations. Meanwhile, the imperfect Refer-Claim Consistency scores indicate that even when relevant references are retrieved, models may not always use them faithfully to support the generated claims. Together, these results reveal a critical weakness: such systems can produce seemingly well-supported answers while relying on questionable evidence or misaligning claims with their cited sources. Therefore, future Search Agent systems should incorporate explicit citation verification mechanisms, such as source credibility assessment, cross-reference validation, and factual reliability checking, to ensure the quality and accuracy of cited evidence. In addition, citation-claim consistency verification mechanism are needed to determine whether each generated claim is genuinely entailed by its corresponding sources, thereby ensuring that references are not only accurate but also used appropriately.

\subsection{Judgment Process}
 
\subsubsection{Instruction Following} 
Instruction following evaluates whether the response correctly executes each decomposed subtask according to its instruction-following rubric. 
The judge model checks whether the response understands the required action, covers the requested content, and satisfies key constraints such as scope, quantity, format, and completeness. 
For example, if a subtask asks for a list of entities within a specified scope, the judge checks whether such a list is provided and whether the scope is respected. The judge model assigns a score from $\{0, 0.5, 1\}$:
\begin{itemize}
    \item \textbf{1 (Fully satisfied):} The subtask is fully satisfied, with all essential requirements and constraints correctly followed.
    \item \textbf{0.5 (Partially satisfied):} The subtask is partially satisfied, but some non-critical requirements are missing or imperfectly handled. For example, when the user requests the top-10 movies of the year with their directors, the agent returns all ten titles but omits director information for some entries.
    \item \textbf{0 (Not satisfied):} The subtask is not satisfied, such as when the response omits the required content, answers irrelevantly, refuses without reason, or fails to perform the required action.
\end{itemize}

\subsubsection{Factuality}
Factuality evaluates whether the objective claims in the response are factually correct. 
For each response report, we extract factual claims based on the factual rubrics. Specifically, the extracted claims must be objective, specific statements verifiable through factual sources. We then construct search queries for each extracted claim and employ an orchestrated workflow equipped with web search (Serper Search\cite{serperdev}) and web fetch (Jina Reader\cite{jinareader}) to verify their correctness. The factuality score of each subtask is quantified as the proportion of verified correct claims among all extracted claims: 
\begin{equation}
    \mathrm{fac}_i = \frac{|\mathcal{C}^{\mathrm{correct}}_i|}{|\mathcal{C}_i|},
\end{equation}
where $\mathcal{C}_i$ denotes the set of factual claims extracted for the $i$-th subtask, and $\mathcal{C}^{\mathrm{correct}}_i$ denotes the subset of claims verified as correct. Furthermore, if the report provides references for factual claims, the corresponding web pages serve as key sources and are jointly considered with other retrieved sources to determine claim correctness. In this process, we additionally measure the information consistency between claims and their cited references, reflecting the tested Search Agent's ability to synthesize information from retrieved web pages.


\subsubsection{Rationality}
Rationality evaluates whether the analytical parts of the response are logically sound and well supported. For each response report, we extract the parts related to the rationality rubrics, which typically involve explanations, comparisons, causal analysis, trade-off evaluation, or recommendations, while excluding factual claims already used for factuality verification to ensure independence between evaluation dimensions. The judge model then assesses whether each extracted parts presents a coherent and reasonable line of reasoning, such as whether the conclusion follows from the stated evidence, whether the comparison criteria are appropriate, and whether the analysis avoids obvious logical gaps or unsupported leaps. The judge model assigns a rationality score from $\{0,0.5,1\}$: a score of 1 indicates that the analysis is coherent, well justified, and directly supports the subtask requirement; a score of 0.5 indicates that the analysis is partially reasonable but contains minor logical gaps, insufficient support, or incomplete discussion; and a score of 0 indicates that the analysis is largely unreasonable, unsupported, irrelevant, or logically flawed.


\subsection{LLM Configuration}

\subsubsection{Deep Research Agents}
For native deep research models, specialized configurations were implemented to accommodate their unique operational characteristics:

\begin{itemize}
    \item \textbf{Autonomous Research Execution}: These models possess fully integrated web search and content synthesis capabilities that operate independently without requiring external tool definitions. The models autonomously determine search strategies, execute queries, retrieve and analyze web content, and synthesize findings into coherent reports. 
    
    \item \textbf{Processing Duration Allowance}: Given the substantially longer execution times inherent to deep research operations, which involve multiple rounds of autonomous web exploration and content synthesis, timeout thresholds were extended to 1,800 seconds. 
    
\end{itemize}

\subsubsection{LLMs with Web Search Tools}
For standard LLMs with external web search tools, the following unified configurations were applied to ensure a standardized evaluation environment:

\begin{itemize}
    \item \textbf{External Tools}: Two external tools were provided to facilitate web-based information retrieval. The \texttt{google\_search} tool enables models to query search engines with custom keywords and retrieve structured organic results containing titles, URLs, and snippets. The \texttt{fetch\_webpage} tool allows models to extract full-text content from any specified URL, primarily utilizing the Jina Reader API for Markdown conversion.
    
    \item \textbf{Extended Reasoning Activation}: To ensure sufficient analytical depth, we enabled the corresponding extended thinking or reasoning features for all models when available. For models supporting the \texttt{thinking} parameter, the thinking budget was set to 8,000 tokens to provide enough capacity for complex multi-step reasoning. For GPT 5.4, the \texttt{reasoning\_effort} parameter was set to \texttt{"medium"}. Kimi-K2.5 has thinking mode enabled by default and thus requires no additional configuration.    
 
    \item \textbf{Response Generation Limits}: The maximum output length was set to 32,768 tokens for all models, ensuring sufficient capacity for generating comprehensive research reports. The temperature was fixed at 1.0 to balance response diversity and reproducibility across repeated evaluations.

  
    \item \textbf{Citation Formatting Protocol}: To support consistent downstream factual verification, all models were instructed to use standardized bracketed numerical citations, such as \texttt{[1][2]}, placed at the end of sentences. Each report was also required to include a unified "References" section at the end, listing all cited sources with their titles and URLs. 
\end{itemize}

\subsubsection{LLMs with Claude Code}

For experiments employing Claude Code as the orchestrating agentic framework with various backend LLMs, the following parameters were established to ensure consistent evaluation:

\begin{itemize}
    \item \textbf{Extended Reasoning Activation}: All backend models integrated within the Claude Code framework were configured with their extended thinking features enabled, following identical parameter settings as described for LLMs with search tools. This ensured that the reasoning capabilities of backend models were fully utilized during the agentic research process, regardless of the orchestration layer.

    \item \textbf{Tool Ecosystem}: MCP (Model Context Protocol) integrations were enabled to provide the systems with comprehensive web research capabilities. Serper was configured as the primary web search provider, offering structured search results with titles, URLs, and snippets. Jina Reader was integrated for webpage content extraction, converting HTML pages to clean Markdown format suitable for LLM consumption. These search tools operated in conjunction with Claude Code's native file system and code execution capabilities.
    
    \item \textbf{Independence}: Session persistence was disabled via the \texttt{--no-session-persistence} flag, ensuring that each question was evaluated independently without contextual carryover from prior tasks. This configuration prevents performance from benefiting from accumulated session knowledge.
    
    \item \textbf{Citation Formatting Protocol}: The same citation normalization procedure as described for standalone LLMs with tool-calling capabilities was applied to all reports generated through the Claude Code framework. This ensured consistent citation structure across different experimental configurations and enabled uniform downstream factual verification using standardized evaluation frameworks.
\end{itemize}

\section{Prompt Templates}
\onecolumn
\begin{tcolorbox} [
    enhanced,
    breakable, 
    colback=black!3,
    colframe=black!65,
    title=System Prompt for Report Generation,
    fonttitle=\bfseries
]
\small
You are a search assistant with web search and webpage reading capabilities who can generate daily report.

\vspace{0.5em}
\textbf{CRITICAL RULES --- you MUST follow all of them:}
\begin{enumerate}
    \item \textbf{NEVER use your own parametric knowledge.} Every factual claim, data point, statistic, name, date, or opinion in your report MUST come from information retrieved via the tools. If you cannot find information through the tools, say so --- do NOT fill in from memory.
    \item \textbf{Research strategy --- fully autonomous, multi-angle verification:} You have complete freedom to decide your research strategy. Use as many search and fetch rounds as needed to produce the most comprehensive, in-depth, and well-verified report possible.
    \item \textbf{Output format --- produce a Markdown research report:}
    \begin{itemize}
        \item Use clear Markdown headings (\#\#, \#\#\#) to organize by topic.
        \item The report should be thorough and at least 2000 words.
        \item Write in the same language as the research question.
    \end{itemize}
    \item \textbf{Citations --- numbered parenthetical references:}
    \begin{itemize}
        \item Assign each source a sequential number starting from 1.
        \item In the report body, cite sources using \textbf{parenthetical numbers}: [1], [2], [3]. For example: ``The three major platforms invested a cumulative 80-100 billion yuan in subsidies [63]''.
        \item Each major claim must be backed by at least one citation.
        \item End with a ``\#\# References'' section listing all cited sources as: \\
        \texttt{[n] Source title\_Site name URL} \\
        For example: \\
        \texttt{[1] Farewell to the cash-burning era! Food delivery platforms simultaneously halt zero-dollar purchases\_Financial News http://example.com/article1} \\
        \texttt{[2] China's fitness industry report 2025\_Reuters https://example.com/article2}
    \end{itemize}
\end{enumerate}

\vspace{0.5em}
When you have gathered sufficient information and are ready, output the final report as your response (without any tool calls). That signals the end of the research process.

\end{tcolorbox}

\begin{tcolorbox}[
    enhanced,
    breakable,
    colback=brown!4!white!96!black,
    colframe=brown!40!black!70,
    title=Instruction Follow Score Prompt,
    fonttitle=\bfseries
    
]
\small
\textbf{1. Role \& Goal} \\
Background time: The current date is \{\texttt{cur\_date}\}. If the question explicitly includes time constraints, please follow the question's requirements.

You are an expert evaluating the ability of an intelligent agent to handle specified tasks. Your focus is on the agent's instruction-following capability. Your scoring must be objective and fair.

\vspace{0.5em}
\textbf{2. Input Format} \\
You will receive the user question (\texttt{Question}), the agent's processing result (\texttt{Document}), and detailed scoring criteria (\texttt{Criteria}) for this evaluation:
\begin{itemize}
    \item \texttt{Question}(str): <User question>
    \item \texttt{Document}(str): <Agent's processing result>
    \item \texttt{Criteria}(list): <Scoring criteria>
\end{itemize}

\vspace{0.5em}
\textbf{3. Workflow} \\
Please strictly follow the workflow below to complete the task:
\begin{enumerate}
    \item Carefully read the \texttt{Question}, \texttt{Document}, and \texttt{Criteria} content. Clearly understand the meaning of each scoring criterion. Do not omit or alter any content in the \texttt{Document}.
    \item Iterate through \texttt{Criteria} and score each individual criterion. Do not add, remove, or modify any scoring criteria. Follow the scoring process below:
    \begin{itemize}
        \item If the \texttt{Document} content strictly satisfies the "criterion" content, the score for this criterion is 1.0.
        \item If the constraints in the "criterion" include multiple subjects, objects, or methods, and only part of them are satisfied in the \texttt{Document}, the score for this criterion is 0.5.
        \item If the \texttt{Document} contains content related to the "criterion", but all content conflicts with the requirements (does not match), the score for this criterion is 0.0.
    \end{itemize}
    \item Carefully verify your scoring result for each criterion to ensure accuracy.
\end{enumerate}

\vspace{0.5em}
\textbf{4. Caution}
\begin{enumerate}
    \item \textbf{Do not judge the accuracy of \texttt{Document} content based on your existing knowledge}. You need to judge based on what the \texttt{Document} claims, even if the content may be incorrect. You do not need to verify its accuracy.
    \begin{itemize}
        \item If the scoring criteria require explicitly providing a certain metric, and the delivery document explicitly states "this metric cannot be obtained", consider the scoring criteria as satisfied.
    \end{itemize}
    \item \textbf{Do not judge based on your time system}. When \texttt{Criteria} contains scoring criteria involving time requirements, please judge based on the time information claimed in the \texttt{Document}.
    \item \textbf{Do not engage in open-ended thinking}. Strictly score according to the standards in \texttt{Criteria} against the \texttt{Document}. Do not add, remove, or alter any standards in \texttt{Criteria}.
    \item \textbf{Your scoring should be very strict}, reflected in the following aspects:
    \begin{enumerate}
        \item All subjects and objects required in the scoring criteria, as well as any actions or conditions related to subjects and objects, must be checked.
        \item Scoring cannot rely solely on section titles in the \texttt{Document}. Verify whether the body text actually contains relevant content that satisfies the scoring criteria.
        \item Body content must explicitly satisfy scoring criteria requirements. Self-inference is prohibited. For example, if the scoring criterion is "Does the delivery document analyze future development based on existing policies?", the body text must explicitly satisfy "analyze future development based on existing policies". Describing "existing policies" and "future development" separately is incorrect.
    \end{enumerate}
    \item \textbf{Ignore the reference materials section}.
\end{enumerate}

\vspace{0.5em}
\textbf{5. Output Format} \\
You must output your scoring results in the following format:

\vspace{0.5em}
{\ttfamily\scriptsize\raggedright
\noindent [ \\
\hspace*{4mm}\{ \\
\hspace*{8mm}"criterion": "$<$Individual scoring criterion, consistent with input$>$,", \\
\hspace*{8mm}"score": $<$Final score for this criterion$>$, \\
\hspace*{8mm}"explain": "$<$Thinking process, strictly consistent with the final score$>$" \\
\hspace*{4mm}\}, \\
\hspace*{4mm}...and more... \\
] \par}

\vspace{0.5em}
\textbf{Please begin your work:} \\
\texttt{Question}: \{question\} \\
\texttt{Document}: \{document\} \\
\texttt{Criteria}: \{criteria\}

\end{tcolorbox}

\begin{tcolorbox} [
    enhanced,
    breakable, 
    colback=cyan!4!white!96!black,
    colframe=cyan!40!black!70,
    title=Claims Extract Prompt,
    fonttitle=\bfseries
]
\small

\textbf{1. Role and Objective} \\
You are a text information mining expert, skilled at locating and extracting ``claim information'' from documents.

\vspace{0.5em}
\textbf{2. Input Format}
\begin{itemize}
    \item \texttt{Document} (str): Input document
    \item \texttt{Question} (str): Accuracy question
\end{itemize}

\begin{quote}
\textbf{Important Principle: All extracted content must originate from the \texttt{Document}, and only claim information related to the \texttt{Question} should be extracted.}
\end{quote}

\vspace{0.5em}
\textbf{3. Workflow}

\textbf{Step 1: Analysis and Clarification}
\begin{itemize}
    \item \textbf{Objective}: Accurately understand the input information.
    \item \textbf{Action}: Deeply analyze the \texttt{Document} and \texttt{Question} to identify all information related to the accuracy question.
    \item \textbf{Note}:
    \begin{enumerate}
        \item ``Delivery result'' in the \texttt{Question} refers to the \texttt{Document}.
        \item Pay attention to headings of all levels in the \texttt{Document}; some headings may directly correspond to the content of the \texttt{Question}.
    \end{enumerate}
\end{itemize}

\textbf{Step 2: Location and Extraction}
\begin{itemize}
    \item \textbf{Objective}: Precisely locate and extract the target information.
    \item \textbf{Action}:
    \begin{enumerate}
        \item Locate the target information and fill the original complete content into the ``fact'' field. \textbf{Modifying the original text in any way is strictly prohibited}.
        \item Integrate the sentences from the ``fact'' content and store them in the ``extract'' field as the final extraction result. Sentence integration is allowed, such as clarifying the objects referred to by pronouns, providing textual interpretations of chart content, supplementing missing background context, etc. However, \textbf{tampering with, adding, or deleting core content is strictly prohibited}, and \textbf{phrases like ``in the delivery result'', ``in the document'', or ``according to the document'' are strictly prohibited}.
    \end{enumerate}
    \item \textbf{Note}:
    \begin{enumerate}
        \item The extracted information must be a \textbf{factual claim}, i.e., an objective, specific statement whose authenticity can be verified through authoritative sources. Subjective evaluations, basic common sense, symbolic metaphors, suggestions/instructions, hypothetical reasoning, and other vague statements that cannot be objectively verified must be excluded.
        \item The extracted information must explicitly appear in the \texttt{Document}. \textbf{Fabricating content that does not exist in the \texttt{Document} is strictly prohibited}.
        \item Extract all relevant content from the \texttt{Document} to avoid any omissions.
        \item If the target information in the \texttt{Document} appears in the form of ``no relevant content support'' or ``no data'', \textbf{it must also be extracted}.
        \item When extracting, \textbf{sufficient context and background information must be supplemented} to avoid semantic incompleteness or ambiguity caused by taking things out of context. Relevant context may be distributed in different parts of the \texttt{Document}; please read through carefully and supplement it.
        \begin{itemize}
            \item Example: When extracting movie starring information, it should be ``The starring actor of [Movie Name] is [Actor Name]'', rather than just ``[Actor Name]''.
            \item Example: When extracting voice compatibility information, it should be ``The AI Voice APP is limited to Huawei phones, other Android phones are not supported'', rather than just ``The AI Voice APP is limited to Huawei phones'' (the latter might be misunderstood as focusing on ``phones'' rather than the ``Huawei brand'').
        \end{itemize}
        \item If the \texttt{Question} involves quantity requirements, ensure the extracted content meets that quantity.
        \item \textbf{If subject, time, or location information is involved, it must be accurately supplemented.}
        \item \textbf{The ``extract'' field must not contain any subjective content}, including subjective judgments, additional explanations, etc.
    \end{enumerate}
\end{itemize}

\textbf{Step 3: Check and Integration}
\begin{itemize}
    \item \textbf{Objective}: Verify whether the complete workflow meets all the notes and output the final result.
    \item \textbf{Action}:
    \begin{enumerate}
        \item Check item by item whether each field meets the requirements.
        \item Integrate the results into a strict JSON object as the content of ``json\_output'':
    \end{enumerate}
\end{itemize}

{\ttfamily\scriptsize\raggedright
\hspace*{14mm}[ \\
\hspace*{16mm}\{ \\
\hspace*{20mm}"fact": $<$Original target text in the Document$>$, \\
\hspace*{20mm}"extract": $<$Integrated target information$>$ \\
\hspace*{16mm}\}, \\
\hspace*{16mm}... \\
\hspace*{14mm}] \par}

\begin{itemize}
    \item[] \textbf{Note}: Even if there are no extraction results, this question must not be skipped; simply output an empty list in ``json\_output''.
\end{itemize}

\vspace{0.5em}
\textbf{4. Output Format} \\
Please output strictly in the following format:

\vspace{0.5em}
\texttt{$<$analysis$>$} \\
Your analysis process \\
\texttt{$<$/analysis$>$} \\
\texttt{$<$json\_output$>$} \\
The extracted results \\
\texttt{$<$/json\_output$>$}

\vspace{0.5em}
\textbf{5. Notes}
\begin{enumerate}
    \item The \texttt{Document} may be a structured Markdown document; please pay attention to heading level symbols (e.g., ``\#\#\#'').
    \begin{itemize}
        \item If a section heading directly corresponds to the \texttt{Question}, \textbf{you must focus on the content of that section to avoid omissions}.
        \item Section headings may contain subject information; necessary subject context should be supplemented during extraction.
    \end{itemize}
    \item \textbf{Be sure to ensure the comprehensiveness of the extraction}, double-check repeatedly, and do not omit anything.
    \item Compared to ``fact'', \textbf{``extract'' is strictly prohibited from losing any original text information}.
    \item \textbf{Claim information must guarantee atomicity}; each claim should contain only one independent, verifiable factual point. Specific splitting rules:
    \begin{itemize}
        \item \textbf{Body paragraphs}: If a paragraph contains multiple independent factual statements (e.g., data of different subjects, information of different dimensions), it must be split into multiple claims; if multiple sentences jointly describe the same factual point, they should be merged into one.
        \item \textbf{Table data}: Split using the \textbf{cell} as the smallest unit; the fact in each cell should be an independent claim (row/column headings need to be supplemented as context). Treating an entire row or the entire table as a single claim is prohibited.
        \item \textbf{Lists/Enumerations}: The fact in each list item should be an independent claim.
    \end{itemize}
\end{enumerate}

\vspace{0.8em}
\vspace{0.3em}
\textbf{Appendix: Core Principles and Exclusion List}

\begin{itemize}
    \item \textbf{Core Definition}: A claim is an objective, specific statement about physical reality or historical records, the authenticity of which can be explicitly verified through authoritative sources.
    \item \textbf{Strictly exclude the following content}:
    \begin{enumerate}
        \item \textbf{Subjective evaluations and descriptions} (e.g., ``gentle'', ``romantic'', ``more native'', ``the soup base is incredibly delicious'')
        \item \textbf{Basic common sense} (e.g., ``the sun rises in the east and sets in the west'', ``Meituan is a platform with a transaction system'')
        \item \textbf{Interpretations, symbols, and metaphors} (e.g., using A as a metaphor for B)
        \item \textbf{Inferences of motives, intentions, and purposes} (e.g., ``in order to...'', ``aimed at...'')
        \item \textbf{Analysis, summaries, and causal inferences} (e.g., ``therefore...'', ``this reflects...'')
        \item \textbf{Suggestions, instructions, and imperatives} (e.g., ``follow this guide'', ``look for Xiangshan Market'', ``don't buy silverware in the ancient city'')
        \item \textbf{Simulations, hypotheses, and self-reasoned results} (e.g., ``assuming a 40\% penetration rate in tier-1 cities and a uniform 50\% savings replacement rate'', ``mathematical modeling of the above scenario is as follows'')
        \item \textbf{Vague statements that cannot be objectively verified} (e.g., ``100 times quieter than daytime'', ``budget travelers can also experience a premium feel'')
        \item \textbf{Descriptions regarding reference links} (e.g., ``the author of paper [1] is Anthony'')
        \item \textbf{Descriptions related only to the document itself} (e.g., ``this report is based on operating data from January 2020 to October 2025'')
    \end{enumerate}
\end{itemize}

\vspace{0.8em}

\textbf{Please begin your work based on the input information:} \\
\texttt{Document}: \{document\} \\
\texttt{Question}: \{question\}

\end{tcolorbox}

\begin{tcolorbox} [
    enhanced,
    breakable, 
    colback=black!3,
    colframe=black!65,
    title=Claims Integrate Prompt,
    fonttitle=\bfseries
]
\small

\textbf{1. Task Objective} \\
Based on the complete document and the extracted claim information, \textbf{deduplicate} and \textbf{reassign} the claims so that each claim belongs to the most matching accuracy question.

\vspace{0.5em}
\textbf{2. Input Format}
\begin{itemize}
    \item \texttt{Document}: \{document\}
    \item \texttt{Assertions}: \{assertions\}
\end{itemize}

\noindent \texttt{Assertions} is a dictionary structure, where the key is the accuracy question, and the value is the list of claim information already assigned under that question. Each claim comes with a unique ``id''.

\vspace{0.5em}
\textbf{3. Workflow}

\textbf{Step 1: Deduplication} \\
Identify duplicate claims on a \textbf{global scale} (across questions + within the same question):
\begin{itemize}
    \item \textbf{Exact Duplicates}: If the ``extract'' of two claims conveys the same fact (even if worded differently), they are considered duplicates. Keep any one of them and delete the rest.
    \item \textbf{Inclusion Relationship}: The core fact of one claim is completely covered by another. In this case, judge: if the refined one already fully contains the target factual information, keep the refined version (more conducive to subsequent item-by-item verification) and delete the verbose version; if the refined version loses key information, keep the more complete one.
    \item \textbf{Non-duplicate Situations (Deletion Prohibited)}:
    \begin{itemize}
        \item The same fact comes from body description vs. table reference $\rightarrow$ \textbf{Does not constitute a duplicate}, keep both.
        \item Two claims involve the same subject but have different focuses $\rightarrow$ \textbf{Does not constitute a duplicate}.
        \item Two claims contain different specific data or details $\rightarrow$ \textbf{Does not constitute a duplicate}.
    \end{itemize}
\end{itemize}

\textbf{Step 2: Reassignment} \\
Adjust the claims to the most matching accuracy question:
\begin{itemize}
    \item Combine the actual position and semantics of the claim in the \texttt{Document} to determine its most matching accuracy question.
    \item \textbf{Prioritize assigning claims to specific scoring rubric questions to avoid piling them up in the fallback category}.
    \item \textbf{Each claim belongs to only one most matching accuracy question}; the same claim is prohibited from appearing in multiple questions. When a claim is related to multiple questions, assign it to the question with a \textbf{narrower, more specific scope}.
\end{itemize}

\textbf{Step 3: Self-Check} \\
After completing deduplication and reassignment, perform the following checks \textbf{item by item}:
\begin{enumerate}
    \item \textbf{Cross-question Uniqueness}: Iterate through the claim IDs under all questions to confirm that no ID appears in two or more questions. If duplicates are found, keep it only under the most matching question.
    \item \textbf{Semantic Deduplication}: Confirm that there are no two claims conveying the same fact (even if worded differently). If found, delete the redundant one.
    \item \textbf{Accidental Deletion Check}: Confirm that each claim in the ``delete'' list is indeed a duplicate, rather than just ``content-related''.
\end{enumerate}

\vspace{0.5em}
\textbf{4. Output Format} \\
Output a strict JSON object:

\vspace{0.5em}
{\ttfamily\scriptsize\raggedright
\noindent \{ \\
\hspace*{4mm}"delete": [List of deleted claim IDs], \\
\hspace*{4mm}"new\_claim": \{Accuracy question: [List of claim IDs under this question], ...\} \\
\} \par}

\vspace{0.5em}
\textbf{5. Notes}
\begin{enumerate}
    \item Pay special attention to headings of all levels in the \texttt{Document} to assist in the reassignment of claim information.
    \item Omitting any accuracy question is prohibited, and the original order of the accuracy questions must be maintained.
    \item \textbf{Deduplication must be conservative}: Only delete claims whose ``extract'' content is truly duplicated or completely included. Claims that are content-related but have different information must be kept. If unsure whether it is a duplicate, choose to keep it.
    \item The output must only use the claim ``id'' for reference; modifying any accuracy question or the original text of the claims is prohibited.
    \item \textbf{Core Principle: Categorize claims into specific accuracy questions as much as possible, avoiding piling them up in the fallback category.}
\end{enumerate}

\end{tcolorbox}

\begin{tcolorbox} [
    enhanced,
    breakable, 
    colback=brown!4!white!96!black,
    colframe=brown!40!black!70,
    title=Query Generate Prompt,
    fonttitle=\bfseries
]
\small

\textbf{1. Role and Objective} \\
You are a web information retrieval expert, skilled at writing query statements for search engine verification based on claim information.

\vspace{0.5em}
\textbf{2. Input Format}
\begin{itemize}
    \item \texttt{Question} (str): The complete question asked by the user to the AI assistant
    \item \texttt{Sub-Question} (str): The sub-question (scoring rubric) split from the \texttt{Question}
    \item \texttt{Assertions} (list): A list of claim information extracted from the AI assistant's reply and related to the current \texttt{Sub-Question}, formatted as follows:
\end{itemize}

\vspace{0.3em}
{\ttfamily\scriptsize\raggedright
\hspace*{10mm} [ \\
\hspace*{14mm}\{ \\
\hspace*{18mm}"fact": $<$Original claim in the Document$>$, \\
\hspace*{18mm}"extract": $<$Standardized factual claim$>$ \\
\hspace*{14mm}\}, \\
\hspace*{14mm}... \\
\hspace*{10mm}] \par}
\vspace{0.3em}

\begin{quote}
\textbf{Important Principle: All operations are strictly targeted at the claim information in \texttt{Assertions}. Generating other claims or query statements on your own is prohibited.}
\end{quote}

\vspace{0.5em}
\textbf{3. Workflow}

\textbf{Step 1: Claim Verification}
\begin{itemize}
    \item \textbf{Objective} : Ensure the claim information is complete and prepare for query generation.
    \item \textbf{Action} : Iterate through each claim in \texttt{Assertions}:
    \begin{enumerate}
        \item If ``extract'' omits the core context information of ``fact'', leading to taking things out of context or ambiguity, supplement and correct it.
        \item Judge whether this claim is \textbf{an exact duplicate} of other claims (i.e., conveys the same core fact) — if it is a duplicate, remove this claim.
    \end{enumerate}
    \item \textbf{Important}: \textbf{Removing claims on the grounds of ``cannot be verified via the internet'' is strictly prohibited.} The input claims have already undergone preliminary screening; this step is only for supplementary correction and deduplication, not for verifiability filtering. For generalized claims, the specific factual components within them should be dismantled in subsequent steps to generate queries.
\end{itemize}

\textbf{Step 2: Identification and Decomposition}
\begin{itemize}
    \item \textbf{Objective}: Prepare for query statement generation.
    \item \textbf{Action}:
    \begin{enumerate}
        \item \textbf{Identification}: Analyze the claim and identify the core information necessary to distinguish its authenticity.
        \item \textbf{Decomposition}: If the claim requires multi-stage, multi-angle verification, further decompose it to facilitate the generation of progressive query statements.
    \end{enumerate}
\end{itemize}

\textbf{Step 3: Generation and Verification}
\begin{itemize}
    \item \textbf{Objective}: Generate high-quality query statements.
    \item \textbf{Action}:
    \begin{enumerate}
        \item \textbf{Statement Generation}: Iterate through the decomposed claims and generate query statements one by one. Each query statement is an independent dictionary structure, where the \textbf{``id'' field is a 0-based index}, and the ``query'' field is the main body of the query statement (required to be a \textbf{yes/no question format}). If the current query depends on the results of other queries, fill in the list of dependent IDs in the ``dependence'' field.
        \item \textbf{Authenticity Verification}: Perform a final verification on each query statement —
        \begin{quote}
        \textbf{Can this query statement be explicitly compared with a recognized objective fact (such as a specific location, institution name, number, geographical location, scientific common sense, etc.) via a search engine to determine its authenticity?}
        \end{quote}
        \begin{itemize}
            \item If ``Yes'', and the query statement is consistent with the information conveyed by the corresponding claim, keep the query statement.
            \item If ``Yes'', but the query statement tampers with or distorts the corresponding claim, it \textbf{must be modified}.
            \item If ``No'', \textbf{this query statement} can be removed, but \textbf{removing the entire claim because of this is strictly prohibited}. Verifiable factual components must be dismantled from the claim to generate at least one query statement for each claim.
        \end{itemize}
    \end{enumerate}
    \item \textbf{Note}:
    \begin{enumerate}
        \item The query statement must be a \textbf{yes/no question format} to support precise and efficient retrieval. If multiple progressive queries are needed, please strictly follow the steps above.
        \item If there is an indirect relationship between the core demand of the \texttt{Sub-Question} and the current claim, please set up progressive query statements through a multi-hop approach.
        \item \textbf{Each query statement must accurately convey the core demand in the \texttt{Sub-Question}; tampering with the intent is strictly prohibited.}
        \item \textbf{Be sure to distinguish the affirmative/negative voice of the claim to avoid semantic reversal.}
        \item The factual content involved in the query statement \textbf{must strictly appear in the current ``extract''; tampering with, adding, or deleting any modifying words and the factual content itself is strictly prohibited.} Generating the current query statement by referencing other ``extract'' content is prohibited.
        \item Remove redundant content unrelated to factual information (e.g., ``according to reliable sources'', ``according to merchant feedback''), but relevant information involving explicit subjects must be retained.
        \item \textbf{Be sure to pay attention to limiting information such as time and location}, as this information is crucial for web retrieval.
    \end{enumerate}
\end{itemize}

\textbf{Step 4: Check and Integration}
\begin{itemize}
    \item \textbf{Objective}: Verify whether the workflow meets all requirements and output the final result.
    \item \textbf{Action}:
    \begin{enumerate}
        \item Check the information completeness of the query statements item by item to ensure \textbf{no omissions, no tampering, and no fabrication} of any content in the claims.
        \item Check the generation quality of the query statements to ensure there are no issues such as ambiguity or unclear semantics.
        \item Integrate the results into a strict JSON object as the content of ``json\_output''.
    \end{enumerate}
\end{itemize}

\vspace{0.5em}
\textbf{4. Output Format} \\
Please output strictly in the following format:

\vspace{0.5em}
\texttt{$<$analysis$>$} \\
Your analysis process \\
\texttt{$<$/analysis$>$} \\
\texttt{$<$json\_output$>$} \\
{\ttfamily\scriptsize\raggedright
\noindent [ \\
\hspace*{4mm}\{ \\
\hspace*{8mm}"fact": $<$Original information$>$, \\
\hspace*{8mm}"extract": $<$Standardized factual claim$>$, \\
\hspace*{8mm}"queries": [ \\
\hspace*{12mm}\{ \\
\hspace*{16mm}"id": $<$Query statement id$>$, \\
\hspace*{16mm}"query": $<$Query statement generated based on the claim$>$, \\
\hspace*{16mm}"dependence": $<$List of ids the query statement depends on$>$ \\
\hspace*{12mm}\}, \\
\hspace*{12mm}... \\
\hspace*{8mm}] \\
\hspace*{4mm}\}, \\
\hspace*{4mm}... \\
] \par}
\texttt{$<$/json\_output$>$}

\vspace{0.5em}
\textbf{5. Notes}
\begin{enumerate}
    \item \textbf{Be sure to ensure the information completeness of the query statements}, ensuring that the query statements \textbf{do not omit, tamper with, or fabricate} any content in the claims, and perform query verification on all content that may involve authenticity.
    \item \textbf{Be sure to ensure the generation quality of the query statements}, preventing the query statements themselves from having issues such as ambiguity or unclear semantics.
    \item \textbf{Be sure to combine the focus of the \texttt{Sub-Question}}; the content required to be queried is only the part of the claim that corresponds to answering the \texttt{Sub-Question}, and it may not be necessary to query the complete claim information.
    \item \textbf{Pay attention to limiting information such as time and space}, please do not omit them.
\end{enumerate}

\vspace{0.8em}

\textbf{Please begin your work:} \\
\texttt{Question}: \{question\} \\
\texttt{Sub-Question}: \{criterion\} \\
\texttt{Assertions}: \{assertions\}

\end{tcolorbox}

\begin{tcolorbox} [
    enhanced,
    breakable, 
    colback=cyan!4!white!96!black,
    colframe=cyan!40!black!70,
    title=Paragraph Extract Prompt,
    fonttitle=\bfseries
]
\small

\textbf{1. Role \& Goal} \\
You are a text information mining expert, skilled at locating and extracting ``paragraph text'' related to specified content from a complete document. \\
Your task is to locate and extract the relevant paragraphs corresponding to the question list (\texttt{Questions}) based on the input document (\texttt{Document}).

\vspace{0.5em}
\textbf{2. Input Format} \\
You will receive the following inputs:
\begin{itemize}
    \item \texttt{Document}(str): The input document
    \item \texttt{Raw Task}(str): The original question
    \item \texttt{Questions}(list): The question list
    \begin{itemize}
        \item Specific format of \texttt{Questions}: [\texttt{Q1}(str), \texttt{Q2}(str), ...]
    \end{itemize}
\end{itemize}
The original question is just for reference and help you clarify the relevant background of the question list. \\
\textbf{You do not need to answer the content of the question list; you only need to extract the paragraphs related to the question list.}

\vspace{0.5em}
\textbf{3. The Complete Workflow} \\
\begin{enumerate}
    \item \textbf{Step 1}: Read each question in \texttt{Questions} and understand the focus and subject of each question.
    \item \textbf{Step 2}: Read the \texttt{Document} and locate the paragraphs related to each question.
    \item \textbf{Step 3}: Completely extract the paragraphs related to each question. Since there may be multiple relevant paragraphs for each question, store them in a list format.
    \item \textbf{Step 4}: Organize the results and conduct a secondary check to ensure the correctness of the results.
\end{enumerate}

\vspace{0.5em}
\textbf{4. Output Format} \\
\textbf{After completing each \texttt{Qi} in \texttt{Questions}, please output strictly in the following format}, \textbf{and outputting any other content is prohibited.}

\vspace{0.5em}
{\ttfamily\scriptsize\raggedright
\noindent \{ \\
\hspace*{4mm}"Q1": \{ \\
\hspace*{8mm}"analysis"(str): $<$Analysis process for question Q1$>$, \\
\hspace*{8mm}"paragraph"(list): $<$Paragraphs related to question Q1$>$ \\
\hspace*{4mm}\}, \\
\hspace*{4mm}"Q2": \{ \\
\hspace*{8mm}"analysis"(str): $<$Analysis process for question Q2$>$, \\
\hspace*{8mm}"paragraph"(list): $<$Paragraphs related to question Q2$>$ \\
\hspace*{4mm}\}, \\
\hspace*{4mm}...and more... \\
\} \par}

\vspace{0.5em}
\textbf{5. Caution} \\
\begin{enumerate}
    \item You need to ensure the comprehensiveness and completeness of the extracted paragraphs, avoiding taking things out of context. Some noise information in the extracted content is tolerable, but omitting any information is prohibited.
    \item Modifying any expressions in the original text is prohibited.
    \item Different questions may correspond to the same paragraph, which can be extracted repeatedly.
    \item You need to respond to each question \texttt{Qi}, and the keys in the output dictionary should start from ``Q1''.
    \item When filling in ``paragraph'', each element must be a complete text paragraph. \textbf{Splitting a continuous piece of text into multiple short sentences is prohibited}.
\end{enumerate}

\vspace{0.8em}

\textbf{Please start working:} \\
\texttt{Document}: \{document\} \\
\texttt{Raw Task}: \{raw\_task\} \\
\texttt{Questions}: \{questions\}

\end{tcolorbox}

\begin{tcolorbox} [
    enhanced,
    breakable, 
    colback=violet!4!white!96!black,
    colframe=violet!40!black!70,
    title=Claims Exclude Prompt,
    fonttitle=\bfseries
]
\small

You will receive a set of paragraph texts in list format, and a numbered list of claims (in dictionary format). Your task this time is to pick out the claims that have appeared in this paragraph text and return the corresponding serial numbers of the claims.

\vspace{0.5em}
\textbf{Input Format:}
\begin{itemize}
    \item \texttt{Paragraph}(list): $<$Paragraph text$>$
    \item \texttt{Claims}(dict): $<$List of claims, where the key is the serial number and the value is the claim information$>$
\end{itemize}

\vspace{0.5em}
\textbf{Output Format:} \\
A list composed of the serial numbers of the claims, requiring a strict JSON object. \textbf{Outputting any other explanatory content is prohibited}.

\vspace{0.5em}
{\ttfamily\scriptsize\raggedright
\noindent \{ \\
\hspace*{4mm}"ids"(list): $<$A list composed of the serial numbers of the claims$>$ \\
\} \par}

\vspace{0.5em}
\textbf{Caution:}
\begin{enumerate}
    \item Numbers outside the given serial number range are prohibited from appearing in the output.
    \item Omitting any claims that appear in the paragraph is prohibited.
\end{enumerate}

\vspace{0.8em}

\textbf{Please start working:} \\
\texttt{Paragraph}: \{paragraph\} \\
\texttt{Claims}: \{claims\_list\}

\end{tcolorbox}

\begin{tcolorbox} [
    enhanced,
    breakable, 
    colback=black!3,
    colframe=black!65,
    title=Rationality Judge Prompt,
    fonttitle=\bfseries
]
\small

\textbf{Role \& Goal} \\
Background time: It's currently \{\texttt{cur\_date}\}. If the question has specific time constraints, please follow the question's requirements. \\
You are an expert in judging text rationality. You are skilled at combining the complete document content to judge whether the narrative of a specified paragraph is reasonable.

\vspace{0.5em}
\textbf{Input Format:}
\begin{itemize}
    \item \texttt{Document}(str): $<$Complete document content$>$
    \item \texttt{Paragraph}(list): $<$Specified paragraph text, there may be multiple paragraphs$>$
    \item \texttt{Question}(str): $<$Question related to the specified paragraph text$>$
    \item \texttt{Claims}(str): $<$Claims appearing in the specified paragraph text, not included in the scope of rationality verification$>$
\end{itemize}

\noindent I will provide the logical relationship of the input content so that you can better understand this task: \\
\texttt{Document} is the complete document written by the testee. The evaluation expert wants to evaluate the question in \texttt{Question}. Through precise paragraph extraction, the relevant content \texttt{Paragraph} is located. At the same time, there are some claims \texttt{Claims} in \texttt{Paragraph} that have been verified through internet retrieval. \\
Next, your task is to judge the rationality of other descriptions in \texttt{Paragraph} excluding \texttt{Claims}.

\vspace{0.5em}
\textbf{Judgment Logic:} \\
Please refer to the following ideas for rationality verification:
\begin{enumerate}
    \item Locate the position of \texttt{Paragraph} in \texttt{Document}, and carefully read its related context.
    \item As factual information that has been verified as correct, \texttt{Claims} do not have rationality errors. For other content in \texttt{Paragraph}, judge whether the following rationality errors exist:
    \begin{itemize}
        \item \textbf{Contradiction}: There is a direct contradiction between \texttt{Paragraph} and the reasoning or expression of the context.
        \item \textbf{Against common sense}: The expression in \texttt{Paragraph} obviously violates objective common sense, such as giving unusable advice, reasons, etc.
        \item \textbf{Reasoning error}: There are errors in mathematical calculations and reasoning in \texttt{Paragraph}; there are errors in additional reasoning based on existing \texttt{Claims}.
        \item \textbf{Semantic confusion}: The expression in \texttt{Paragraph} has semantic confusion problems, such as secretly changing the subject, grammatical errors, and various faulty wordings. It does not involve punctuation, Emoji, and other related issues.
    \end{itemize}
    \item Your rationality judgment needs to have a solid basis. Judgments with obvious personal preferences and subjectivity are prohibited. At the same time, you do not have the ability to search for objective facts, and please do not judge the authenticity of a claim based on your existing knowledge.
    \item When there are multiple paragraphs of text in \texttt{Paragraph}, you need to make a judgment for each paragraph of text.
    \item Please strictly distinguish the difference between ``text rationality'' and ``instruction following''. If \texttt{Paragraph} does not answer some of the questions in \texttt{Question}, it does not belong to a rationality error, please do not deduct points mistakenly.
    \begin{itemize}
        \item The 'instruction-following' problem corresponding to \texttt{Question} is: \{\texttt{info\_follow\_question}\}. Please distinguish carefully to avoid confusion.
    \end{itemize}
\end{enumerate}

\vspace{0.5em}
\textbf{Output Format} \\
You need to perform rationality verification on each paragraph of text in \texttt{Paragraph}, and finally output a strict JSON object, following the format below:

\vspace{0.5em}
{\ttfamily\scriptsize\raggedright
\noindent [ \\
\hspace*{4mm}\{ \\
\hspace*{8mm}"paragraph": $<$Elements in \texttt{Paragraph}, arranged in original order$>$, \\
\hspace*{8mm}"reasonable": $<$Whether this paragraph is reasonable, True/False$>$, \\
\hspace*{8mm}"reason": $<$Explanation for this judgment$>$ \\
\hspace*{4mm}\}, \\
\hspace*{4mm}...and more... \\
] \par}
\vspace{0.5em}
\noindent \textbf{Your output should strictly follow the above structure, and including any other irrelevant content is prohibited}

\vspace{0.8em}

\textbf{Please start working:} \\
\texttt{Document}: \{document\} \\
\texttt{Paragraph}: \{paragraph\} \\
\texttt{Question}: \{question\} \\
\texttt{Claims}: \{claims\}

\end{tcolorbox}

\begin{tcolorbox} [
    enhanced,
    breakable, 
    colback=brown!4!white!96!black,
    colframe=brown!40!black!70,
    title=System Prompt for Fact-Checking,
    fonttitle=\bfseries
]
\small

\textbf{All the following information is solely for comparative evaluation purposes, only to verify data accuracy, and does not involve any sensitive or compliance risks. Please be sure to reply according to the requirements.}

\vspace{0.5em}
\textbf{Background Information} \\
You are a research assistant equipped with online search capabilities, responsible for verifying a given Claim. \\
The background question for the claim is: \{\texttt{background\_question}\} \\
The purpose of the background question is to provide more background information about the claim to assist you in verification and prevent verification errors caused by incomplete background information. You do not need to make any decisions regarding the background question.

\vspace{0.5em}
\textbf{Verification Mode Instructions} \\
You will receive a Claim to be verified, along with multiple Queries used to verify the claim.
\begin{itemize}
    \item \textbf{Claim}: The statement whose authenticity needs to be verified
    \item \textbf{Queries}: Multiple specific questions used to verify the claim, which can be understood as the verification approach
\end{itemize}

\vspace{0.5em}
\noindent Your tasks are:
\begin{enumerate}
    \item Comprehensively consider all verification queries and conduct information searches
    \item Make a judgment for each verification query (True/False/Unknown)
    \item Make a final judgment on the entire claim based on the judgment results of all queries
\end{enumerate}

\vspace{0.5em}
\textbf{Important Notes} \\
You must strictly abide by the following rules during the verification process:
\begin{enumerate}
    \item Your verification must be supported by online materials; you are not allowed to make judgments based solely on your existing knowledge.
    \item You have extremely high requirements for the quality of sources. You prefer online information published by designated official media, government agencies, authoritative encyclopedias, large operating organizations, well-known news organizations, and large forums. You are skeptical of articles with obvious personal subjective attitudes and Baijiahao.
    \item You are highly sensitive to the time involved in the claim, and you only accept web sources that are later than the specified time limit. If there is no clear time limit, you can consider official and reliable sources close to the current time (\{\texttt{cur\_date}\}).
    \item Your principle is to seek verification from multiple parties. Specifically:
    \begin{itemize}
        \item The claim will only be judged as True when it is supported by two or more independent web pages. Similarly, it will only be judged as False when two or more independent web pages contradict the claim.
        \item If it is difficult to find two or more independent web pages, at least one web page must come from an official agency or authoritative organization to support the decision.
        \item If only one web page is related to the claim and other web pages do not mention it, the conclusion of that web page shall prevail.
        \item Even for the website you consider most authoritative, you should not completely rely on its content. If there are multiple other sub-authoritative websites whose conclusions contradict it, you should discard the viewpoint of that authoritative website.
    \end{itemize}
    \item You need to make a judgment for each verification query separately, and ultimately make an overall judgment on the claim by synthesizing the judgment results of all queries:
    \begin{itemize}
        \item If all queries are judged as ``True'', the claim is ``True''
        \item If any query is judged as ``False'', the claim is ``False''
        \item If there is an ``Unknown'' and no ``False'', the claim is ``Unknown''
    \end{itemize}
    \item You are a research assistant specializing in the text modality. If the verification relies entirely on multimodal information such as images, audio, or video, please select ``Unknown''.
\end{enumerate}

\end{tcolorbox}

\begin{tcolorbox} [
    enhanced,
    breakable, 
    colback=cyan!4!white!96!black,
    colframe=cyan!40!black!70,
    title=Next Action Selection for Search Prompt,
    fonttitle=\bfseries
]
\small

\textbf{Context} \\
You need to verify a Claim this time. Below is the claim and its related list of verification queries. Search History is used to record your previous search and summary history.

\vspace{0.3em}
\textbf{Claim to be verified}: \{claim\}

\vspace{0.3em}
\textbf{Verification approach (Queries)}: Below are multiple query questions used to verify this claim. You need to comprehensively consider these queries to search for information: \\
\{queries\}

\vspace{0.3em}
\textbf{Search History}: \{context\}

\vspace{0.5em}
\textbf{Action Space} \\
\vspace{0.3em}
[1] search \\
\hspace*{4mm}Description: Retrieve query-related information on the Internet \\
\hspace*{4mm}Parameters: \\
\hspace*{8mm}- query (str): The query statement used for retrieval

\vspace{0.3em}
[2] answer \\
\hspace*{4mm}Description: Make an overall judgment on the claim based on current existing knowledge. \\
\hspace*{4mm}Parameters: \\
\hspace*{8mm}- reference (str): The reference materials used for the final answer

\vspace{0.5em}
\textbf{Next Action} \\
Please select the next action to be executed based on all existing information (including the claim to be verified, verification approach, search history) and the action space. \\
Specifically, if you believe the existing information is sufficient to answer all verification queries and make an overall judgment on the claim, you will select ``answer'' as the next action; otherwise, you will select ``search'' for further searching. \\
Please note that when you select ``search'' for further searching, you need to fill in the specific query statement used for the search. The writing rules are as follows:
\begin{enumerate}
    \item The format should be a complete short sentence, not just a few keywords.
    \item The query statement \textbf{must not lose key information from the verification queries}, such as time, location, person, event, method, etc. Avoid causing ambiguity.
    \item You can refer to the query list in the verification approach to construct the search statement, or you can optimize it yourself based on the actual situation.
\end{enumerate}

\vspace{0.5em}
\textbf{Output Format} \\
\textbf{Please strictly output the following JSON object, and any other characters are prohibited.}

\vspace{0.5em}
{\ttfamily\scriptsize\raggedright
\noindent \{ \\
\hspace*{4mm}"action": $<$Next action, search or answer$>$, \\
\hspace*{4mm}"reason": $<$Reason for selecting this action$>$, \\
\hspace*{4mm}"search\_query": $<$If the next action is "search", fill in the specific query statement used for the search; if the next action is "answer", return an empty string$>$ \\
\} \par}
\texttt{"""}

\end{tcolorbox}

\begin{tcolorbox} [
    enhanced,
    breakable, 
    colback=black!3,
    colframe=black!65,
    title=Webpage Content Verification Prompt,
    fonttitle=\bfseries
]
\small

The Claim you need to verify this time is: \{claim\} \\
The related verification queries are: \{queries\}

\vspace{0.5em}
\textbf{Subtask Description} \\
You have currently obtained a potentially relevant webpage, and you need to judge whether this webpage can help answer the above verification queries, thereby verifying the claim. \\
Below is some information about this relevant webpage:
\begin{itemize}
    \item \textbf{Title}: \{page\_title\}
    \item \textbf{Content}: \{page\_content\}
    \item \textbf{Date}: \{page\_date\}
\end{itemize}

\vspace{0.5em}
\textbf{Workflow}
\begin{enumerate}
    \item Carefully read the webpage content information provided above, and examine the true degree of relevance of this information to the claim and each verification query.
    \item For each verification query, judge whether this webpage can provide an answer.
    \item When the information in the webpage is sufficient to make a decision (True or False) for a certain verification query, you need to return the relevant partial information and the decision result. If you think the webpage information is irrelevant to a certain query, return None in the corresponding field.
    \item After completing the above tasks, integrate all results into \textbf{a strict JSON object}.
\end{enumerate}

\vspace{0.5em}
\textbf{Special Attention Examples}
\begin{enumerate}
    \item You are very sensitive to numbers. When the number of digits is the same, any discrepancy will be considered an error. When the number of digits is different, you can accept rounding. For ranges represented by numbers, you uphold the same standard as the former. For example:
    \begin{itemize}
        \item 100 and 110 have the same number of digits and do not accept rounding, so they will be considered an error.
        \item 100.1 and 100 have a different number of digits and meet the rounding standard, so they will be considered correct.
    \end{itemize}
    \item It is prohibited to add any subjective inference or extension during the judgment process. All judgments must be made based on the verification queries and the actual webpage content.
\end{enumerate}

\vspace{0.5em}
\textbf{Output Format} \\
\textbf{Please strictly output the following JSON object, and any other characters are prohibited.}

\vspace{0.5em}
{\ttfamily\scriptsize\raggedright
\noindent \{ \\
\hspace*{4mm}"relevant\_context": $<$Key information extracted from the webpage related to the verification, which needs to be sufficient and have enough context information$>$, \\
\hspace*{4mm}"query\_results": [ \\
\hspace*{8mm}\{ \\
\hspace*{12mm}"query\_id"(int): $<$The ID of the verification query$>$, \\
\hspace*{12mm}"query": $<$The content of the verification query$>$, \\
\hspace*{12mm}"flag": $<$One of True/False/None, indicating the judgment of this webpage on this query$>$, \\
\hspace*{12mm}"evidence": $<$Specific evidence supporting this judgment, fill in null if flag is None$>$ \\
\hspace*{8mm}\}, \\
\hspace*{8mm}...provide results for each verification query... \\
\hspace*{4mm}], \\
\hspace*{4mm}"explanation": $<$Your overall explanation for this analysis$>$ \\
\} \par}

\end{tcolorbox}

\begin{tcolorbox} [
    enhanced,
    breakable, 
    colback=brown!4!white!96!black,
    colframe=brown!40!black!70,
    title=Claim Aggregation and Judgment Prompt,
    fonttitle=\bfseries
]
\small

\textbf{Claim to be verified}: \{claim\} \\
\textbf{Extraction source}: \{extract\} \\
\textbf{Verification approach}: \\
\{queries\}

\vspace{0.5em}
\textbf{Subtask Description} \\
You have currently obtained some relevant webpage materials and completed the preliminary verification. You need to:
\begin{enumerate}
    \item Provide an answer for each verification query
    \item Make a final judgment on the entire claim based on the answers to all queries
\end{enumerate}

\vspace{0.3em}
Field descriptions:
\begin{itemize}
    \item ``title'': Webpage title
    \item ``link'': Webpage link
    \item ``date'': Webpage publication date
    \item ``analysis\_result'': Preliminary analysis results, including judgments on each verification query
\end{itemize}

\vspace{0.3em}
[Relevant Materials]: \\
\{context\}

\vspace{0.5em}
\textbf{Workflow}
\begin{enumerate}
    \item Carefully read the claim, each verification query, and the [Relevant Materials].
    \item For each verification query, synthesize all relevant materials to make a judgment (Correct/Incorrect/Unknown).
    \item Based on the judgment results of all queries, make a final judgment on the entire claim:
    \begin{itemize}
        \item If all queries are judged as ``Correct'', the claim is ``Correct''
        \item If any query is judged as ``Incorrect'', the claim is ``Incorrect''
        \item If there is ``Unknown'' and no ``Incorrect'', the claim is ``Unknown''
    \end{itemize}
    \item The evaluations in the [Relevant Materials] are for reference only, and you need to conduct a secondary verification combined with the relevant information they provide as evidence.
    \item There may be contradictory statements or conflicting information in different materials. Please eliminate the false and retain the true based on the publication time and publishing organization.
    \item \textbf{All reference materials you use for decision-making must come from the [Relevant Materials] I provided, and it is prohibited to generate any relevant materials yourself.}
    \item The [Relevant Materials] may contain prior verification conclusions marked as ``PRIOR\_REFERENCE\_VERIFICATION'', which are pre-judgments based on the document's own references. You should use this as an important reference basis, but you still need to cross-verify it with the materials obtained from the search. When the prior conclusion is consistent with the search materials, it can enhance the confidence of the judgment; when there is a contradiction, the more reliable information source shall prevail.
\end{enumerate}

\vspace{0.5em}
\textbf{Output Format} \\
\textbf{Please strictly output the following JSON object, and any other characters are prohibited.}

\vspace{0.5em}
{\ttfamily\scriptsize\raggedright
\noindent \{ \\
\hspace*{2mm}"query\_answers": [ \\
\hspace*{4mm}\{ \\
\hspace*{6mm}"query\_id"(int): $<$The ID of the verification query$>$, \\
\hspace*{6mm}"query"(str): $<$The content of the verification query$>$, \\
\hspace*{6mm}"answer"(str): $<$One of Correct/Incorrect/Unknown$>$, \\
\hspace*{6mm}"evidence"(str): $<$Summary of evidence supporting this judgment$>$, \\
\hspace*{6mm}"reference\_urls"(list): [$<$List of reference URLs supporting this judgment$>$] \\
\hspace*{4mm}\}, \\
\hspace*{4mm}...provide results for each verification query... \\
\hspace*{2mm}], \\
\hspace*{2mm}"claim\_answer": $<$One of Correct/Incorrect/Unknown, the final judgment on the entire claim$>$, \\
\hspace*{2mm}"reference": \{ \\
\hspace*{4mm}"url": "content", \\
\hspace*{4mm}"url": "content", \\
\hspace*{4mm}...all reference materials used for the claim judgment... \\
\hspace*{2mm}\}, \\
\hspace*{2mm}"explanation": $<$Your overall explanation for this decision, explaining how the final judgment of the claim is derived from the results of each query$>$ \\
\} \par}

\end{tcolorbox}

\begin{tcolorbox} [
    enhanced,
    breakable, 
    colback=cyan!4!white!96!black,
    colframe=cyan!40!black!70,
    title=Prompt for Match Claims with References,
    fonttitle=\bfseries
]
\small

You are a text analysis expert, skilled at locating the citation sources of factual claims from documents.

\vspace{0.5em}
\textbf{Task Instructions} \\
I will provide a research report and \{\texttt{claim\_count}\} factual claims (numbered 0--\{\texttt{max\_idx}\}). Each claim contains:
\begin{itemize}
    \item \textbf{fact}: The factual claim extracted from the report (may be slightly rewritten)
    \item \textbf{extract}: The originally extracted text snippet
\end{itemize}

\noindent Please complete the following tasks for each claim:

\vspace{0.3em}
\textbf{1. Locate the Original Text}
\begin{itemize}
    \item Use the fact and extract as clues to search for the corresponding original text location in the document
    \item The original text may have slight differences from the fact/extract (such as formatting or punctuation), but the core content should be consistent
\end{itemize}

\vspace{0.3em}
\textbf{2. Extract Citations} \\
After locating the original text, extract all relevant citation markers. Pay attention to the context of the original text; a single fact may correspond to multiple citations, do not miss any.

\vspace{0.5em}
\textbf{Citation Format Instructions} \\
Citations in the main body of the research report may appear in the following formats, please identify them carefully:
\begin{enumerate}
    \item \textbf{A piece of text + [number]} (there may be multiple) \\
    For example: ``divided society into 7 classes[15]'' or ``[15][16]''
    \item \textbf{A piece of text + [(number)](link)} \\
    For example: ``divided society into 7 classes[(15)](https://www.example.com)''
    \item \textbf{[citation:N]} format \\
    For example: ``Lefit's single stores usually achieve break-even within a few months of opening [citation:53]''
    \item \textbf{} or \textbf{} format \\
    For example: ``only about 4\%--5\% in China''
\end{enumerate}

\vspace{0.5em}
\textbf{Extraction Notes}
\begin{enumerate}
    \item \textbf{Ensure original\_text is complete and understandable}: It must be a complete and understandable sentence or paragraph
    \item \textbf{Handling multiple citations}: If there are multiple citations, list all of them in the references array
    \item \textbf{ref\_idx extraction rules}: Only extract the number part, ignore position information (e.g., for [15$\dagger$L10], only extract ``15'')
    \item \textbf{Reference URL lookup}: Look up the URL based on the index from the reference list at the end of the document
    \item \textbf{No citation case}: If the original text has no citation markers, references should be an empty array
\end{enumerate}

\vspace{0.5em}
\textbf{Output Format} \\
Please directly output a JSON array, where each element corresponds to a claim:

\vspace{0.5em}
{\ttfamily\scriptsize\raggedright
\noindent [ \\
\hspace*{2mm}\{ \\
\hspace*{4mm}"located": true, \\
\hspace*{4mm}"original\_text": "Original text snippet containing citation markers", \\
\hspace*{4mm}"has\_reference": true, \\
\hspace*{4mm}"references": [ \\
\hspace*{6mm}\{"ref\_idx": "1", "ref\_url": "https://example1.com"\}, \\
\hspace*{6mm}\{"ref\_idx": "2", "ref\_url": "https://example2.com"\} \\
\hspace*{4mm}] \\
\hspace*{2mm}\}, \\
\hspace*{2mm}\{ \\
\hspace*{4mm}"located": false, \\
\hspace*{4mm}"original\_text": null, \\
\hspace*{4mm}"has\_reference": false, \\
\hspace*{4mm}"references": [] \\
\hspace*{2mm}\} \\
] \par}

\vspace{0.5em}
Field descriptions:
\begin{itemize}
    \item \texttt{located}: boolean, whether the original text was found in the document
    \item \texttt{original\_text}: string or null, the original text content (including citation markers)
    \item \texttt{has\_reference}: boolean, whether there is a citation
    \item \texttt{references}: array, each element contains ref\_idx (string) and ref\_url (string or null)
\end{itemize}

\vspace{0.8em}

\textbf{Please begin your work:} \\
\textbf{Document}: \\
\{document\}

\vspace{0.5em}
\textbf{Claims} (Total of \{claim\_count\}): \\
\{claims\}

\vspace{0.5em}
Please directly output a JSON array, you must return \{claim\_count\} elements, do not output any explanations.

\end{tcolorbox}

\begin{tcolorbox} [
    enhanced,
    breakable, 
    colback=violet!4!white!96!black,
    colframe=violet!40!black!70,
    title=Prompt for Reference Consistency Judgment,
    fonttitle=\bfseries
]
\small

You will see a reference material and some statements. Please judge whether the statement is support, conflict, or unknown with respect to the reference material. Note:

\vspace{0.5em}
\textbf{Judgment Criteria:}
\begin{enumerate}
    \item First, determine whether the reference material contains valid content. If there is no valid information in the reference material (such as a ``page not found'' page, garbled text, or irrelevant content), then the status of all statements is considered \textbf{unknown}.
    \item If the reference material is valid:
    \begin{itemize}
        \item \textbf{support}: The facts or data contained in the statement can be fully or partially found in the reference material (data rounding is acceptable)
        \item \textbf{conflict}: The facts or data contained in the statement explicitly contradict or conflict with the content in the reference material
        \item \textbf{unknown}: The relevant information of the statement is neither supported nor contradicted in the reference material, making it impossible to judge
    \end{itemize}
\end{enumerate}

\vspace{0.5em}
\textbf{Output Format:} \\
Return a JSON list, where each item contains:
\begin{itemize}
    \item \texttt{idx}: $<$The sequence number of the statement$>$
    \item \texttt{result}: $<$The judgment result (support/conflict/unknown)$>$
    \item \texttt{explanation}: $<$Relevant explanation for the reason of the judgment$>$
    \item \texttt{context}: $<$Relevant evidence information extracted from the reference material (extract the supporting original text for support, extract the conflicting original text for conflict, leave as an empty string for unknown)$>$
\end{itemize}

\vspace{0.5em}
For example:

\vspace{0.3em}
{\ttfamily\scriptsize\raggedright
\noindent [ \\
\hspace*{4mm}\{ \\
\hspace*{8mm}"idx": 1, \\
\hspace*{8mm}"result": "support", \\
\hspace*{8mm}"explanation": "The reference material explicitly mentions this data", \\
\hspace*{8mm}"context": "According to reports, the company's revenue reached 50 million yuan in 2024..." \\
\hspace*{4mm}\}, \\
\hspace*{4mm}\{ \\
\hspace*{8mm}"idx": 2, \\
\hspace*{8mm}"result": "conflict", \\
\hspace*{8mm}"explanation": "The data shown in the reference material does not match the statement", \\
\hspace*{8mm}"context": "The report shows that the company's revenue in 2024 was only 30 million yuan..." \\
\hspace*{4mm}\}, \\
\hspace*{4mm}\{ \\
\hspace*{8mm}"idx": 3, \\
\hspace*{8mm}"result": "unknown", \\
\hspace*{8mm}"explanation": "The reference material does not mention relevant information", \\
\hspace*{8mm}"context": "" \\
\hspace*{4mm} \\
] \par}
\vspace{0.5em}

Below are the reference material and statements: \\
\texttt{$<$reference$>$} \\
\{reference\} \\
\texttt{$<$/reference$>$}

\vspace{0.5em}
\texttt{$<$statements$>$} \\
\{statements\} \\
\texttt{$<$/statements$>$}

\vspace{0.5em}
Begin the judgment below, directly output the JSON list, do not output any chit-chat or explanations.

\end{tcolorbox}

\end{document}